%% file: main.tex
\pdfoutput=1

\documentclass{article}

\usepackage[preprint]{neurips_2026}

\usepackage[utf8]{inputenc}
\usepackage[T1]{fontenc}
\usepackage[breaklinks=true]{hyperref}
\usepackage{url}
\expandafter\def\expandafter\UrlBreaks\expandafter{\UrlBreaks\do\-}
\usepackage{booktabs}
\usepackage{array}
\usepackage{amsfonts}
\usepackage{amsmath}
\usepackage{amssymb}
\usepackage{xcolor}
\usepackage{graphicx}
\usepackage{rotating}
\usepackage{tikz}

\graphicspath{{figures/}}

\title{GB-LSR: Local Spectral Decoding with a Learned Global Bandwidth
       for Arbitrary-Scale Super-Resolution\thanks{Code:
       \href{https://github.com/KempnerInstitute/gblsr}{\nolinkurl{github.com/KempnerInstitute/gblsr}}\quad
       Models: \href{https://huggingface.co/KempnerInstituteAI/gblsr}{\nolinkurl{huggingface.co/KempnerInstituteAI/gblsr}}}}

\author{%
  Max Shad\thanks{Corresponding author.}\\
  Kempner Institute for the Study of Natural\\
  and Artificial Intelligence\\
  Harvard University\\
  \texttt{max\_shad@harvard.edu}%
  \And
  Naeem Khoshnevis\\
  Kempner Institute for the Study of Natural\\
  and Artificial Intelligence\\
  Harvard University\\
  \texttt{naeem\_khoshnevis@harvard.edu}%
}

\begin{document}

\maketitle

\input{sections/abstract.tex}

\input{sections/01_introduction.tex}
\input{sections/02_related_work.tex}
\input{sections/03_method.tex}
\input{sections/04_experiments.tex}
\input{sections/05_results.tex}

\input{sections/07_limitations.tex}
\input{sections/08_conclusion.tex}

\section*{Acknowledgments}
This work has been made possible in part by a gift from the Chan
Zuckerberg Initiative Foundation to establish the Kempner Institute
for the Study of Natural and Artificial Intelligence at Harvard
University.

\bibliographystyle{plainnat}
\bibliography{references}

\appendix
\input{sections/appendix.tex}

\end{document}

%% file: sections/abstract.tex
\begin{abstract}
We present GB-LSR (Global-Bandwidth Local Spectral Representation),
a fixed-grid local spectral representation for continuous image
decoding. The image domain is partitioned into non-overlapping
square patches. Each patch carries coefficients for a truncated
Fourier basis, predicted by a single linear projection from shared
convolutional-encoder features, and one trainable scalar bandwidth
is shared across every patch and every image. As in earlier local
spectral decoders, decoding at a continuous coordinate is a fixed-size
basis contraction whose cost is set by the spectral cutoff; GB-LSR
learns the bandwidth of that basis instead of fixing it.
We evaluate an arbitrary-scale super-resolution extension,
GB-LSR-Scalar-ASR, against the authors' released LIIF, LTE, and SRNO
checkpoints on the same RDN encoder, with every method scored under one
protocol and timed in one session per scale, each on one GPU. It runs
$1.25\times$ faster than LIIF-RDN at $\times 4$ and as fast as SRNO-RDN, whose released code uses
$15$ times as much peak memory on Urban100. It trails the three encoder-matched
baselines by $0.07$ to $0.79$\,dB PSNR-Y in distribution, SRNO-RDN by
$0.35$\,dB on average. Removing the local ensemble raises the speedup
to $2.41\times$ over LIIF-RDN and $1.95\times$ over SRNO-RDN at $\times 4$,
and to $3.00\times$ and $2.41\times$ at $\times 8$, without changing PSNR-Y
beyond seed variation, at the cost of value jumps at
cell boundaries of $0.22$ gray levels (of $255$) on average at $\times 4$.
Against the EDSR-baseline checkpoints of five recent
methods at $\times 4$, GB-LSR-Scalar-ASR scores above or within
$0.17$\,dB on PSNR-Y of LMF, SRNO-EDSR, and OPE-SR-EDSR ($1.39$ to
$6.43$ million parameters against $22.02$) and $0.14$ to $0.57$\,dB
below GSASR and Thera ($20.44$ and $5.85$ million), and has a higher mean
LPIPS at $\times 4$ than every baseline.
\end{abstract}

%% file: sections/01_introduction.tex
\section{Introduction}
\label{sec:introduction}

Continuous image representations let one store an image as a function
over its continuous coordinate domain rather than a fixed pixel grid,
and then query any coordinate at any density at inference time.
Multi-layer perceptrons (MLPs) with Fourier-feature inputs, sinusoidal
activations, or wavelet activations are fit per
image~\citep{tancik2020fourier,sitzmann2020siren,saragadam2023wire},
while local implicit functions amortize the field across a
distribution: LIIF~\citep{chen2021liif} decodes each query with a ReLU
MLP on encoder features and local coordinates, and
LTE~\citep{lee2022lte} adds Fourier embeddings of those coordinates.
Arbitrary-scale extensions have explored attention-based and
neural-operator decoders (CiaoSR, \citealp{cao2023ciaosr}; CLIT,
\citealp{chen2023clit}; SRNO, \citealp{wei2023srno}; SSRNO,
\citealp{han2024ssrno}), decoding in a latent space (LMF,
\citealp{he2024lmf}), decoding without a learned network through
orthogonal position encoding (OPE-SR, \citealp{song2023opesr}), and
iterative feature-diffusion layers with cross-attention (DIIN,
\citealp{dai2025diin}). Two properties matter in practice:
expressiveness, meaning how accurately the representation can
reconstruct the image, and inference cost, meaning how many operations
a single pixel query needs. Local spectral decoders such as OPE-SR
address the second: they store coefficient tensors over a fixed
spectral basis on a grid, so each query touches a constant-size
neighborhood and costs a fixed number of operations.

We present GB-LSR (Global-Bandwidth Local Spectral
Representation), a fixed-grid local spectral representation in
which a single trainable scalar bandwidth, which sets the frequency of
the basis, is shared across every patch in every image. The image
domain is partitioned into a fixed grid of non-overlapping square
patches; each patch carries a small block of coefficients for a
truncated Fourier basis, predicted from shared convolutional-encoder
features by a single linear projection. Reconstruction at any
continuous query coordinate is a fixed-size
basis contraction whose cost is independent of image size, so the
encoder is paid once per image while the decoder pays
$O(p_\text{max}^2)$ multiply-adds per output coordinate, where
$p_\text{max}$ is the number of modes per axis of the truncated
Fourier basis
(Section~\ref{sec:method:basis}).

We evaluate GB-LSR on arbitrary-scale super-resolution, the standard
benchmark for this family of methods, against the released checkpoints
of LIIF, LTE, and SRNO on the RDN encoder, of LTE on SwinIR, of OPE-SR on a
$147$-channel RDN, and of five recent methods on EDSR-baseline encoders,
all scored under one protocol and timed in one session per scale. A
design study on native reconstruction compares GB-LSR variants with one
another (Section~\ref{sec:experiments}).

\paragraph{Contributions.}
\begin{enumerate}
  \item We propose GB-LSR, a fixed-grid local spectral representation
    that learns one scalar bandwidth shared across every patch and
    image, where OPE-SR fixes the frequency of its basis and Thera
    learns a global bank of 2D frequencies; like OPE-SR, it decodes a
    coordinate with a fixed number of multiply-adds set by the spectral
    cutoff (Sections~\ref{sec:related} and~\ref{sec:method}). In the
    design study, learning the bandwidth from a poor start gains $3.7$ to
    $5.5$\,dB over holding it there
    (Appendix~\ref{app:adaptive_bw_ablation}); in super-resolution it
    stays near its start, and starting it at $2.0$ instead of $1.0$ costs
    $2.2$\,dB in distribution (Table~\ref{tab:asr_sensitivity}).
  \item On arbitrary-scale super-resolution, GB-LSR-Scalar-ASR, the
    extension of Section~\ref{sec:method:asr}, runs $1.25\times$ faster
    than LIIF-RDN and $1.21\times$ faster than LTE-RDN at $\times 4$ and
    as fast as SRNO-RDN, whose released code uses $15$ times as much
    peak memory on Urban100, all on the same RDN encoder. It trails the three by $0.07$ to
    $0.79$\,dB in peak signal-to-noise ratio on the luminance channel
    (PSNR-Y) across the in-distribution scales and test sets, SRNO-RDN by
    $0.35$\,dB on average. Removing the local ensemble raises the speedup
    to $2.41\times$ over LIIF-RDN and $1.95\times$ over SRNO-RDN at
    $\times 4$, and to $3.00\times$ and $2.41\times$ at $\times 8$, with
    PSNR-Y unchanged within seed variation, at the cost
    of value jumps at cell boundaries of $0.22$ gray levels (of $255$) on
    average at $\times 4$. Perceptual quality is its
    weakest axis (Section~\ref{sec:results}).
  \item In the design study, at a fixed spectral cutoff and feature
    width, halving the native patch size from $32$ to $16$ improves whole-image
    PSNR-RGB by $6.13$\,/\,$5.66$\,/\,$5.55$\,dB on
    Kodak\,/\,Set14\,/\,Urban100 with $14.9\%$ fewer parameters but four
    times the per-image latent ($8{,}192$ to $32{,}768$ values); at a
    matched latent of $16{,}384$ values and $p_\text{max} = 16$ the two
    patch sizes are within
    $0.14$\,dB in PSNR-RGB, although the $P = 32$ model then has $15.8$
    times the parameters (Appendix~\ref{app:patchsize}).
  \item We test whether the bandwidth should vary from patch to patch
    and find no clear gain from a per-patch field over one global scalar in
    the design study (Section~\ref{sec:lim:locality},
    Appendix~\ref{app:design}).
\end{enumerate}

%% file: sections/02_related_work.tex
\section{Related work}
\label{sec:related}

\paragraph{Continuous and coordinate-based image representations.}
Coordinate-based neural fields treat an image, or more generally a
signal, as a continuous function over its coordinate domain and fit an
MLP to that function. Two ingredients are central: Fourier-feature input
mappings~\citep{tancik2020fourier}, which address the spectral bias of
plain ReLU MLPs, and sinusoidal hidden-layer
activations~\citep{sitzmann2020siren}, which let coordinate-based MLPs
represent fine spatial detail and signal derivatives. Together these
made coordinate-based fitting practical for natural images.
WIRE~\citep{saragadam2023wire} replaces the sinusoidal activation with a
complex-Gabor wavelet, giving a spectral and localization trade-off
inside the MLP itself.

\paragraph{Local implicit representations.} A second line of work adds a
local component, typically by conditioning an MLP decoder on features
from a grid of image patches. Meta-SR~\citep{hu2019metasr} was the
first to upsample to an arbitrary scale, predicting the weights of its
upsampling layer from the scale factor.
LIIF~\citep[local implicit image function;][]{chen2021liif} predicts
color at a query coordinate from the surrounding encoder-feature cells;
LTE~\citep[local texture estimator;][]{lee2022lte}
uses per-coordinate Fourier embeddings carrying estimated amplitudes and
dominant frequencies.

Later decoders add context or change the query-time model.
A-LIIF~\citep{li2022aliif} makes the local implicit function adaptive,
CLIT~\citep{chen2023clit} adds cross-scale local attention,
HIIF~\citep{jiang2025hiif} builds on a hierarchical encoding, and
LIWT~\citep{duan2024liwt} decodes from wavelet sub-bands.
SRNO~\citep{wei2023srno} casts upsampling as a neural operator, and
GSASR~\citep{chen2025gsasr} renders 2D Gaussians instead of querying an
MLP. Thera~\citep{becker2024thera} evaluates a per-pixel neural heat
field, a one-hidden-layer network over one globally learned bank of 2D
frequencies whose phases and amplitudes a hypernetwork predicts and
whose thermal activation attenuates each component with the output
scale; GB-LSR instead scales a fixed separable harmonic basis with one
learned scalar.

\paragraph{Fixed-grid local spectral bases.} The representation we study
is a fixed-grid local spectral basis with per-patch coefficient tensors:
each patch carries a small set of coefficients over a fixed basis, and
the reconstruction at any query coordinate is a bounded local
combination of those coefficients. OPE-SR~\citep{song2023opesr} decodes
this way without a learned decoder: it contracts one coefficient block
per low-resolution feature cell against a separable truncated Fourier
basis and reports lower decoding time than MLP decoders. GB-LSR also
decodes with a fixed-size contraction but learns the bandwidth, which OPE-SR
fixes at the equivalent of $s = 0.5$ (Section~\ref{sec:method:basis}),
uses $16$ modes per axis rather than $7$, predicts the coefficients with
a linear head on a $64$-channel RDN encoder where OPE-SR widens the
encoder to $147$ channels to emit them directly, and natively places
one block per patch of side $P$ rather than per input pixel. The bandwidth
of the basis is the central hyperparameter, and a natural design
question is whether it should vary spatially across patches or be
shared globally. We test per-patch alternatives in the design study
(Appendices~\ref{app:locality} and~\ref{app:adaptive_bw_ablation}).

\paragraph{Decoding cost.} Other methods also target the cost of
arbitrary-scale decoding. LMF~\citep{he2024lmf} moves most decoder work
onto the low-resolution latent grid and keeps a small MLP per query,
SSRNO~\citep{han2024ssrno} reports about twice the throughput of SRNO in a
compressed configuration,
and GSASR rasterizes its Gaussians with a dedicated GPU kernel. LMF and
SSRNO still evaluate a learned network at each query; OPE-SR and GB-LSR
evaluate none.

\paragraph{Perceptual metrics.} We report the learned perceptual image
patch similarity~\citep[LPIPS-AlexNet;][]{zhang2018lpips} alongside PSNR
and the structural similarity index~\citep[SSIM;][]{wang2004ssim}. LPIPS compares deep features of an
ImageNet-trained AlexNet and correlates better with human perceptual
judgment than PSNR or SSIM alone for natural images.

%% file: sections/03_method.tex
\section{Method}
\label{sec:method}

GB-LSR represents an image as a fixed grid of local spectral coefficient
blocks over a truncated Fourier basis, with a single trainable scalar
bandwidth shared across every patch and every image. The encoder runs
once per image; decoding any continuous coordinate is then a fixed-size
contraction against the basis. This section defines the basis, states
the resulting per-query cost, describes behavior at patch boundaries,
specifies the three ways of handling the bandwidth that we study, and
extends the decoder to arbitrary-scale super-resolution.
Figure~\ref{fig:architecture} shows the path from image to color.

\begin{figure}[t]
  \centering
  \begin{tikzpicture}[
      font=\small,
      box/.style={draw, rounded corners=2pt, minimum height=8mm, align=center, inner sep=3pt},
      ar/.style={-latex, thick},
    ]
    \node[box, minimum width=15mm] (img) at (0,0) {image\\$x$};
    \node[box, minimum width=17mm] (enc) at (2.5,0) {convolutional\\encoder $E$};
    \node[box, minimum width=17mm] (feat) at (5.2,0) {features\\$z = E(x)$};
    \node[box, minimum width=19mm] (proj) at (8.2,0) {linear\\projection};
    \node[box, minimum width=20mm] (coef) at (11.4,0) {coefficients\\$c_e$ per patch};

    \draw[ar] (img) -- (enc);
    \draw[ar] (enc) -- (feat);
    \draw[ar] (feat) -- (proj);
    \draw[ar] (proj) -- (coef);

    \node[box, minimum width=15mm] (query) at (0.9,-2.1) {query $u$};
    \node[box, minimum width=26mm] (local) at (4.7,-2.1)
      {local coordinate $\hat{u}_e$ for\\each patch $e$ whose\\support contains $u$};
    \node[box, minimum width=20mm] (basis) at (9.9,-2.1)
      {basis $\psi(\hat{u}_e; s)$};

    \draw[ar] (query) -- (local);
    \draw[ar] (local) -- (basis);

    \node[box, minimum width=30mm] (out) at (5.6,-3.9)
      {$f(u) = \sum_e \chi_e(u)\, \langle \psi(\hat{u}_e; s),\, c_e \rangle \in \mathbb{R}^3$};

    \draw[ar] ([xshift=6mm]coef.south) |- (out.east);
    \draw[ar] (basis) -- (out);

    \node[anchor=east] (bw) at (9.2,-1.05)
      {bandwidth $s$: one scalar, shared globally};
    \draw[ar, dashed] (bw.east) -| (basis.north);
  \end{tikzpicture}
  \caption{\textbf{GB-LSR decoding path.} The encoder runs once per image
  and a single linear projection produces one coefficient block $c_e$ per
  patch. A query $u$ is mapped to the local coordinate $\hat{u}_e$ of
  each patch whose support contains it, and the decoded color sums, over
  those patches, the Fourier basis evaluated at $\hat{u}_e$ contracted with
  $c_e$, weighted by $\chi_e(u)$. For the native models one patch
  contributes; the super-resolution decoder blends four neighboring
  patches unless trained without the ensemble
  (Section~\ref{sec:method:boundaries}). One trainable scalar bandwidth
  $s$ serves every patch in every image.}
  \label{fig:architecture}
\end{figure}

\subsection{Fixed-grid local spectral basis}
\label{sec:method:basis}

An image $x \in \mathbb{R}^{3 \times H \times W}$ is encoded into a
feature map $z = E(x)$ by a shared convolutional encoder. The image
plane is partitioned into a fixed grid of non-overlapping square patches
of side $P$. A single linear projection maps the encoder feature at each
patch to a coefficient tensor
$c_e \in \mathbb{R}^{3 \times p_\text{max} \times p_\text{max}}$, one
$p_\text{max} \times p_\text{max}$ block per color channel, over a fixed
separable basis with $p_\text{max}$ modes per axis.

Reconstruction at a continuous query coordinate $u$ is
\begin{equation}
  f(u) \;=\; \sum_{e \,:\, u \in \text{supp}(e)}
    \chi_e(u) \, \bigl\langle \psi\bigl(\hat{u}_e; \, s\bigr),
    \, c_e \bigr\rangle,
  \label{eq:local_spectral_eval}
\end{equation}
where $\chi_e$ is the patch's partition-of-unity weight, $u_e$ is the
patch center, and $\hat{u}_e = 2(u - u_e)/P$ is the patch-local
coordinate, which lies in $[-1, 1]^2$ on the patch itself and in
$[-2, 2]^2$ over the overlapping supports of
Section~\ref{sec:method:boundaries}.

\paragraph{The basis.} $\psi(\,\cdot\,; s)$ is separable: entry
$[\psi(v; s)]_{ij} = \varphi_i(v_1)\,\varphi_j(v_2)$ for the 1D mode list
\begin{equation}
  \varphi_0(t) = 1, \qquad
  \varphi_{2j-1}(t) = \cos(j \pi s t), \qquad
  \varphi_{2j}(t) = \sin(j \pi s t), \qquad j = 1, 2, \ldots
  \label{eq:basis_modes}
\end{equation}
taken in this order until exactly $p_\text{max}$ modes exist per axis.
For even $p_\text{max}$ the list ends on an unpaired cosine at
$j = p_\text{max}/2$. The bandwidth $s$ multiplies the angular argument
of every non-constant mode, so it sets the frequency of the whole basis
at once. With $s = 0.5$ and an odd $p_\text{max}$, this basis equals, up
to a factor $\sqrt{2}$ on the non-constant modes, the fixed-frequency
basis that OPE-SR~\citep{song2023opesr} uses with its patch ensemble.
The inner product in \eqref{eq:local_spectral_eval} contracts the two
mode indices per color channel under a smooth per-mode cutoff, so
$f(u) \in \mathbb{R}^3$ with
$[\langle \psi, c_e \rangle]_k = \sum_{i,j} w_{ij} [\psi]_{ij} [c_e]_{kij}$
and $w_{ij} = \sigma\bigl((p_\text{soft} - \max(i, j)) \kappa\bigr)$,
$\sigma$ the logistic sigmoid and $\kappa = 4$ a fixed sharpness. In the
design study, the cutoff order $p_\text{soft}$ is held at $16$ unless
GB-LSR-Full predicts it (Section~\ref{sec:method:variants}), and at $32$
in the one configuration with $p_\text{max} = 32$
(Appendix~\ref{app:patchsize}), so at $p_\text{max} = 16$ the highest
modes keep weight $\sigma(\kappa) \approx 0.98$ and at $p_\text{max} = 12$
every weight is $1$ to within $10^{-8}$.

\subsection{Cost of a query}
\label{sec:method:cost}

The basis is fixed and the partition is local, so a query touches a
constant-size neighborhood of patches whatever the image size.
Evaluating one patch at one coordinate takes $2 p_\text{max}$ scalar
modes and a contraction of about $3 p_\text{max}^2$ multiply-adds. A
query evaluates one patch under the indicator weights of
Section~\ref{sec:method:boundaries}; under its overlapping weights it
evaluates four, taking $8 p_\text{max}$ modes and about
$12 p_\text{max}^2$ multiply-adds plus a four-term weighted average. Its
work is thus $O(p_\text{max}^2)$, set by the spectral cutoff and the
number of blended patches, and independent of $H$, $W$, and the
requested output resolution. The encoder runs once per image, and its
cost is amortized over every query drawn from it; as in OPE-SR, no
learned network is evaluated at a query.

\subsection{Behavior at patch boundaries}
\label{sec:method:boundaries}

The weights $\chi_e$ in \eqref{eq:local_spectral_eval} decide how the
blocks of the non-overlapping grid meet. The native models use indicator
weights,
$\chi_e(u) = \mathbf{1}\{u \in \text{supp}(e)\}$ with each support equal
to its patch, so exactly one term is nonzero. Within a patch the
reconstruction is then smooth, being a finite sum of sinusoids, but
neighboring patches carry separate coefficient blocks, so $f$ may jump
across the boundary between them. The super-resolution decoder
(Section~\ref{sec:method:asr}), whose patches are the cells of the
low-resolution feature grid, instead blends the four cells whose
centers surround the query with bilinear area weights: the local
ensemble of LIIF~\citep{chen2021liif}, which OPE-SR~\citep{song2023opesr}
also uses for rendering. Each weight is supported on the square two
cells wide centered on its cell, vanishes at the edge of that square,
and the weights sum to one, so the output is continuous in value, with
kinks in its derivative along the lines through cell centers. The
super-resolution variants trained without the ensemble use indicator
weights and are piecewise smooth like the native models. The blocks on
either side of a boundary are predicted by a shared convolutional
encoder whose receptive field spans it, so training with indicator
weights can limit the jumps but not remove them;
Appendix~\ref{app:boundary} measures them.

\subsection{Bandwidth handling}
\label{sec:method:variants}

The bandwidth $s$ in \eqref{eq:basis_modes} is the spectral parameter
the model adapts; GB-LSR-Full also adapts the cutoff order. We study
three ways of handling it, all sharing the same encoder and basis.

\begin{description}
  \item[GB-LSR-Scalar.] One global trainable scalar, applied identically
    to every patch in every image and trained end to end with the
    encoder and decoder. In the design study it starts at $s = 0.707$, the
    geometric midpoint of its range $[0.25, 2.0]$, and learns $0.756$ at
    $P = 32$.

  \item[GB-LSR-Fixed.] One global scalar held at $s = 1.125$, the linear
    midpoint of the same range, with the cutoff order held at
    $p_\text{max}$, so no spectral parameter is trained. It therefore
    differs from GB-LSR-Scalar in both the value of $s$ and its training.
    Held instead at the value GB-LSR-Scalar learns, $s$ gives the same
    whole-image PSNR-RGB to within $0.006$\,dB, so the $3.67$ to
    $5.55$\,dB gap between the two comes from the value of $s$
    (Appendix~\ref{app:adaptive_bw_ablation}).

  \item[GB-LSR-Full.] A per-patch bandwidth field $s_e$ predicted by a
    linear head on spatial encoder features, replacing the global scalar
    in \eqref{eq:local_spectral_eval}. The same head predicts a
    per-patch cutoff order; GB-LSR-Bandwidth is the same model with that
    order held at $p_\text{max}$ (Appendix~\ref{app:design}).
\end{description}

\paragraph{Parameterization.} The two settings bound $s$ differently. In
the design study each trainable bandwidth passes through a log-space
sigmoid onto $[0.25, 2.0]$, with values of $s$ given as implemented
(Appendix~\ref{app:design}). In the super-resolution extension the raw
parameter passes through a softplus, so $s$ is strictly positive and
unbounded above, initialized at $s = 1.0$. From this start, the trained
value stays well inside the range of the design study, so the bound is
not active in practice. With the bandwidth started at $0.5$
instead, the super-resolution decoder reaches the same in-distribution
PSNR-Y within seed variation; started at $2.0$, it loses $2.2$\,dB, and
in both cases the learned value stays near its start
(Table~\ref{tab:asr_sensitivity}).

\paragraph{Why one bandwidth may be enough.} A single scalar is a strong
restriction: it asks one frequency scale to serve flat sky and dense
brickwork alike. It is less restrictive than it appears, because the
bandwidth sets where the basis places its modes, not how much energy
each mode carries, and the per-patch coefficients remain free to
concentrate on low modes in smooth regions and spread across high modes
in textured ones. Local content variation can therefore be absorbed by
the coefficients rather than by $s$. What a per-patch bandwidth would add is
the ability to place modes differently in different regions, which
matters only if the useful frequency support varies more across patches
than the coefficients can follow. Appendix~\ref{app:design} tests
whether it does in native reconstruction.

\paragraph{Why a fixed basis rather than a learned decoder.} Replacing
$\psi$ with a learned coordinate network makes each query more
expressive and more costly, since the network is evaluated at each
output coordinate. A fixed basis, as in OPE-SR, moves the learning into
the coefficients, which depend on the patch but not on the query, and
leaves the per-query path as a contraction whose size is known in
advance. The trade is representational freedom at a coordinate against
learned computation that runs once per image rather than once per
coordinate.

\subsection{Arbitrary-scale extension}
\label{sec:method:asr}
For arbitrary-scale super-resolution, the decoder is placed on an RDN
encoder~\citep{zhang2018rdn} operating at the low-resolution grid. A
linear projection maps each low-resolution feature cell to one
coefficient block, so the cell plays the role of the patch and $P$ is
its high-resolution footprint, which depends on the requested scale.
Queries are taken at the target grid and decoded by
\eqref{eq:local_spectral_eval} with the local ensemble of
Section~\ref{sec:method:boundaries} unless stated otherwise. This decoder
applies no per-mode cutoff ($w_{ij} = 1$); a fixed per-mode weight could
be absorbed into its linear projection.

%% file: sections/04_experiments.tex
\section{Experimental setup}
\label{sec:experiments}

We evaluate GB-LSR on two tasks. Arbitrary-scale super-resolution, the
standard benchmark for this family of methods, is where we compare
against other methods. Native reconstruction is used only as a design
study of the representation itself, comparing GB-LSR variants against
one another.

\paragraph{Comparison principles.} Every baseline is a checkpoint
released by its authors and run unmodified, so none is retrained or
resized to meet a parameter budget, and trainable parameter counts are
reported (Table~\ref{tab:asr_quality}) without being forced to match. All
methods share the same low-resolution inputs, ground-truth images, and
metric code, and are timed in the same session at each scale.

\paragraph{Arbitrary-scale super-resolution.} GB-LSR-Scalar-ASR, the
extension of Section~\ref{sec:method:asr}, and its variants train for $1{,}000{,}000$ steps on
DIV2K~\citep{agustsson2017div2k} with three seeds, minimizing an L1 loss on
$48{\times}48$ low-resolution patches with scales sampled from $[1, 4]$, at
batch size~$16$ with random horizontal flips and $90^\circ$ rotations, and
using Adam at a learning rate of $10^{-4}$ halved at $200{,}000$,
$400{,}000$, $600{,}000$, and $800{,}000$ steps. For
LIIF~\citep{chen2021liif}, LTE~\citep{lee2022lte}, and
SRNO~\citep{wei2023srno} we use the authors' released checkpoints with
their released model code and inference routine: LIIF-RDN, LTE-RDN, and
SRNO-RDN share the RDN encoder~\citep{zhang2018rdn} with
GB-LSR-Scalar-ASR, at $22.32$, $22.47$, and $22.77$ against $22.02$
million trainable parameters, and LTE-SwinIR uses the SwinIR
encoder~\citep{liang2021swinir} at $12.10$ million. We also evaluate the
released OPE-SR~\citep{song2023opesr} on a $147$-channel RDN
(OPE-SR-RDN, trained at $\times 4$ only) and five recent methods on
EDSR-baseline encoders~\citep{lim2017edsr}: LMF~\citep{he2024lmf},
SRNO (SRNO-EDSR), OPE-SR with $147$ channels (OPE-SR-EDSR, also
trained at $\times 4$ only),
GSASR~\citep{chen2025gsasr}, and Thera~\citep{becker2024thera} without its
self-ensemble. Of the recent methods we screened, seven had code and
checkpoints that we located: these five, and CiaoSR and SSRNO, which need
additional software builds and were deferred. \mbox{A-LIIF}, CLIT, HIIF, LIWT,
and DIIN are cited but not evaluated, nor are the RDN checkpoints that
LMF, GSASR, and Thera also release (Appendix~\ref{app:asr}). Evaluation covers Set5~\citep{bevilacqua2012set5},
Set14~\citep{zeyde2010set14}, B100~\citep{martin2001bsds},
Urban100~\citep{huang2015urban100}, and the DIV2K validation split, at
in-distribution scales $\times 2$\,/\,$\times 3$\,/\,$\times 4$ and
out-of-distribution scales $\times 6$\,/\,$\times 8$, and, for
GB-LSR-Scalar-ASR (base), LIIF-RDN, LTE-RDN, LTE-SwinIR, and OPE-SR-RDN,
on the DIV2K validation split also at $\times 12$ to $\times 30$; one test
set at one scale is a cell.

\paragraph{Metrics.} Reconstruction quality is PSNR-Y and SSIM-Y on the
BT.601 luminance channel, with a border as wide as the scale removed on
every test set. Perceptual quality is LPIPS-AlexNet~\citep{zhang2018lpips}
on the full RGB output, with one implementation for every method.
Low-resolution inputs are PIL bicubic downsamplings of the ground truth
cropped to a multiple of the scale; the Set5, Set14, B100, and Urban100
ground truth is first cropped at the bottom and right to a multiple of
$12$ pixels, and the DIV2K validation images are not. Because the luminance
convention also applies to the DIV2K validation split, those cells are
comparable across the methods here but not with published DIV2K
validation numbers, which use RGB and a larger border. Given the authors'
inputs, our scoring reproduces $80$ published scores of four baselines
within $0.023$\,dB (Appendix~\ref{app:additional_evidence}).

\paragraph{Timing.} Every method is timed at $\times 4$ in one session on one
NVIDIA H200 GPU, and at $\times 8$ in a second session on another H200,
at batch size~1 without gradients, from the same low-resolution
inputs, with $10$ warm-up and $50$ timed passes per image; a pass covers
the whole inference of one image, coordinate construction included. We
report the median over the timed passes, averaged over the images of a
cell and, for the models we train, over three seeds; speed ratios are
geometric means of per-cell ratios. We also report peak allocated GPU
memory on the most demanding cell. Appendix~\ref{app:asr} gives the
per-method chunking, framework, and timer.

\paragraph{Design study.} The design study trains GB-LSR variants on a
mixture of DTD~\citep{cimpoi2014dtd} and DIV2K for $1{,}000{,}000$
steps at batch size~$8$ over three seeds and compares design choices
such as patch size and bandwidth handling under an otherwise fixed
recipe. Each image is reconstructed from its own latent, $(256/P)^2$
encoder features of width $d_\text{feat}$: $8{,}192$ values at $P = 32$
and $32{,}768$ at $P = 16$ for $d_\text{feat} = 128$, $24$ and $6$ times
fewer than its $196{,}608$ pixel values. Quality is whole-image PSNR-RGB, SSIM, and
LPIPS on held-out Kodak~\citep{kodak}, Set14, and Urban100 standardized
to $256{\times}256$; bandwidth statistics use $256$ unseen validation
images.

%% file: sections/05_results.tex
\section{Arbitrary-scale super-resolution}
\label{sec:results}

\subsection{Quality}
\label{sec:results:asr}

\begin{figure}[t]
  \centering
  \includegraphics[width=\linewidth]{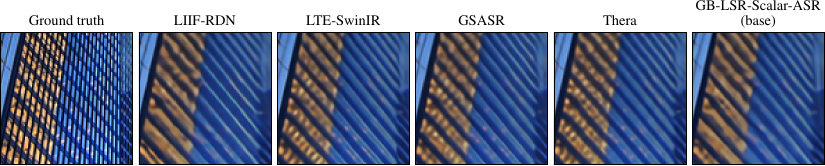}
  \caption{\textbf{Qualitative comparison at $\times 4$.} Urban100
  \texttt{img\_012}, the same $96{\times}96$ window in every tile, chosen
  from the ground truth alone: among windows aligned with the
  low-resolution pixel grid, the one with the largest mean squared
  difference between neighboring luma values. All methods receive the
  same bicubic low-resolution input; the baselines use their authors'
  released checkpoints, Thera without its self-ensemble, and the GB-LSR
  tile shows one of its three trained models. Every method shows diagonal
  stripes where the ground truth has a vertical grid of windows; the
  bicubic input already carries them. Appendix~\ref{app:qualitative} shows every method.}
  \label{fig:qualitative_asr}
\end{figure}

Table~\ref{tab:asr_quality} reports PSNR-Y at $\times 4$ on the five test
sets and mean LPIPS; Appendix~\ref{app:asr} gives the grids from
$\times 2$ to $\times 8$
(Tables~\ref{tab:asr_grid_psnr}--\ref{tab:asr_grid_lpips} and
Figure~\ref{fig:asr_psnr_vs_scale}), and Figure~\ref{fig:qualitative_asr}
shows one image.

\paragraph{Encoder-matched baselines.} LIIF-RDN, LTE-RDN, and SRNO-RDN
lead GB-LSR-Scalar-ASR on every in-distribution cell: LIIF-RDN by
$0.07$ to $0.36$\,dB PSNR-Y, $0.16$\,dB on average, LTE-RDN by $0.11$ to
$0.52$\,dB, $0.24$\,dB on average, and SRNO-RDN by $0.16$ to $0.79$\,dB,
$0.35$\,dB on average,
all three by the most on Urban100 $\times 2$. Out of distribution the
leads shrink to $0.03$ to $0.25$\,dB, $0.07$ to $0.42$\,dB, and $0.10$ to
$0.51$\,dB. At $\times 12$ to $\times 30$ on the DIV2K validation split,
GB-LSR-Scalar-ASR is within $0.03$\,dB of LIIF-RDN at every scale, with
LTE-RDN ahead by at most $0.09$\,dB (Table~\ref{tab:asr_ood_ladder});
SRNO-RDN was not scored at these scales.

\paragraph{Other released baselines.} LTE-SwinIR leads on every
in-distribution cell, by $0.20$ to $0.99$\,dB, and by $0.13$ to
$0.21$\,dB at $\times 12$ to $\times 30$. OPE-SR-RDN, trained at
$\times 4$ only, trails at $\times 2$ by $0.19$ to $0.74$\,dB and leads
at every scale from $\times 4$ to $\times 30$, by $0.05$ to $0.24$\,dB at
$\times 4$ and $0.03$ to $0.04$\,dB at $\times 12$ to $\times 30$; its
encoder width, mode count, coefficient head, and training scales all
differ from those of GB-LSR-Scalar-ASR, so the differences cannot be
assigned to any one of them.

\paragraph{Recent methods.} At $\times 4$, GB-LSR-Scalar-ASR scores above
LMF and OPE-SR-EDSR on all five test sets and within $0.17$\,dB of SRNO-EDSR,
three methods with $1.39$ to $6.43$ million parameters; it scores $0.14$
to $0.57$\,dB below GSASR ($20.44$ million) and Thera ($5.85$ million) on
every test set. Mean LPIPS at $\times 4$ is highest for the three GB-LSR
rows; every baseline scores a lower LPIPS than the base model on
every test set, except OPE-SR-EDSR on Urban100 and DIV2Kval, where the
two differ by at most $0.0014$. LTE-SwinIR and GSASR have the lowest mean
LPIPS.

\input{tables/asr_quality_merged_main.tex}

\subsection{Latency and memory}
\label{sec:results:cost}

Table~\ref{tab:asr_latency} reports latency and peak allocated memory
from the $\times 4$ timing session of Section~\ref{sec:experiments},
and Figure~\ref{fig:asr_quality_latency_x4} in Appendix~\ref{app:asr} plots
Urban100 PSNR-Y against latency.
Against the three encoder-matched baselines, GB-LSR-Scalar-ASR runs
$1.25\times$ faster than LIIF-RDN, $1.21\times$ faster than LTE-RDN, and
within $1\%$ of the speed of SRNO-RDN; it runs $1.12\times$ faster than
OPE-SR-RDN and $4.05\times$ faster than LTE-SwinIR. Of the recent methods
on EDSR-baseline encoders, LMF runs $2.79\times$ faster; SRNO-EDSR,
OPE-SR-EDSR, and Thera run $1.38$ to $1.42\times$ faster; and GSASR runs
$2.75\times$ slower.

\paragraph{Within the GB-LSR family.} Removing the local ensemble
(Section~\ref{sec:method:boundaries}) reduces latency by
$1.93\times$, raising the speedup over LIIF-RDN to
$2.41\times$ and over SRNO-RDN to $1.95\times$; in the $\times 8$ session they
rise to $3.00\times$ and $2.41\times$ (Table~\ref{tab:asr_latency_x8}), with
its $\times 8$ PSNR-Y $0.03$ to $0.12$ and $0.11$ to $0.38$\,dB below LIIF-RDN and SRNO-RDN on
Set14, B100, and Urban100. The base model reaches $1.29\times$ and
$1.03\times$ at $\times 8$, which, given that session's seed spread, we do
not count as a change from $\times 4$.
At cell boundaries, where the base variant is continuous, the output of the variant without the ensemble
jumps by $0.22$ gray levels (of $255$) on average at $\times 4$, and by less than one on every image at that scale
(Appendix~\ref{app:boundary}), yet in-distribution PSNR-Y
moves by only $+0.002$\,dB as a three-seed mean, within the base
variant's seed standard deviation of $0.015$\,dB, with a worst cell of
$-0.008$\,dB on Set14 $\times 4$. Widening the RDN encoder to 96 channels
in the variant without the ensemble moves PSNR-Y relative to the base
variant by $+0.016$\,dB in distribution, about one seed standard
deviation of the base variant, and by $+0.017$\,dB out of distribution, while keeping a
$1.76\times$ reduction (Table~\ref{tab:asr_family}).

\paragraph{Memory.} On Urban100 $\times 4$, GB-LSR-Scalar-ASR peaks at
$899$\,MiB of allocated memory, LIIF-RDN, LTE-RDN, and LTE-SwinIR at
$901$ to $922$\,MiB, OPE-SR-RDN at $2{,}333$\,MiB, and SRNO-RDN at
$13{,}692$\,MiB. Of the recent methods, OPE-SR-EDSR and LMF need less,
$316$ and $590$\,MiB, while SRNO-EDSR and GSASR need $13{,}613$ and
$8{,}077$\,MiB; both SRNO models and GSASR evaluate all queries at once
(Appendix~\ref{app:asr}). Within the GB-LSR family the peak follows the encoder
width, not the local ensemble (Table~\ref{tab:asr_family}).

\input{tables/asr_latency_merged_main.tex}

%% file: tables/asr_quality_merged_main.tex
\begin{table}[t]
  \centering
  \caption{\textbf{Arbitrary-scale super-resolution quality at $\times 4$.} PSNR-Y (dB) on the five test sets and LPIPS averaged over them, both scored as in Section~\ref{sec:experiments}; DIV2Kval is the DIV2K validation set. The GB-LSR-Scalar-ASR rows are means over three training seeds; the other rows are single checkpoints released by their authors. Params: trainable parameters in millions. RDN-96: RDN with $96$ instead of $64$ feature channels; EDSR-b: EDSR-baseline. OPE-SR-RDN uses a $147$-channel RDN (RDN-147) and OPE-SR-EDSR a $147$-channel EDSR-baseline (EDSR-b-147); both were trained at $\times 4$ only. The suffix noLE marks training and evaluation without the local ensemble (Section~\ref{sec:method:boundaries}), and nf96 the RDN-96 encoder. Thera is scored without its self-ensemble. Bold: best value in each quality column.}
  \label{tab:asr_quality}
  \fontsize{8}{9.5}\selectfont
  \setlength{\tabcolsep}{3.5pt}
  \begin{tabular}{llrrrrrrr}
    \toprule
    Method & Encoder & Params & \multicolumn{5}{c}{PSNR-Y at $\times 4$ (dB) $\uparrow$} & Mean \\
    \cmidrule(lr){4-8}
     &  & (M) & Set5 & Set14 & B100 & Urban100 & DIV2Kval & LPIPS $\downarrow$ \\
    \midrule
    GB-LSR-Scalar-ASR (base) & RDN & 22.02 & 32.21 & 28.73 & 27.67 & 26.45 & 30.68 & 0.2750 \\
    GB-LSR-Scalar-ASR-noLE & RDN & 22.02 & 32.21 & 28.73 & 27.67 & 26.46 & 30.68 & 0.2745 \\
    GB-LSR-Scalar-ASR-nf96+noLE & RDN-96 & 24.93 & 32.21 & 28.75 & 27.67 & 26.46 & 30.70 & 0.2755 \\
    \midrule
    LIIF-RDN & RDN & 22.32 & 32.49 & 28.80 & 27.75 & 26.68 & 30.75 & 0.2568 \\
    LTE-RDN & RDN & 22.47 & 32.60 & 28.88 & 27.78 & 26.81 & 30.81 & 0.2560 \\
    SRNO-RDN & RDN & 22.77 & 32.68 & 28.96 & 27.83 & 26.98 & 30.90 & 0.2521 \\
    OPE-SR-RDN & RDN-147 & 29.65 & 32.45 & 28.80 & 27.73 & 26.53 & 30.73 & 0.2601 \\
    \midrule
    LTE-SwinIR & SwinIR & 12.10 & \textbf{32.80} & \textbf{29.05} & \textbf{27.87} & \textbf{27.24} & \textbf{30.98} & \textbf{0.2491} \\
    LMF & EDSR-b & 1.39 & 32.19 & 28.61 & 27.61 & 26.12 & 30.47 & 0.2679 \\
    SRNO-EDSR & EDSR-b & 2.02 & 32.37 & 28.78 & 27.68 & 26.49 & 30.61 & 0.2600 \\
    OPE-SR-EDSR & EDSR-b-147 & 6.43 & 32.16 & 28.66 & 27.60 & 26.13 & 30.50 & 0.2704 \\
    GSASR & EDSR-b & 20.44 & 32.78 & 28.90 & 27.81 & 27.01 & 30.89 & 0.2511 \\
    Thera & EDSR-b & 5.85 & 32.67 & 29.00 & 27.84 & 27.02 & 30.93 & 0.2567 \\
    \bottomrule
  \end{tabular}
\end{table}

%% file: tables/asr_latency_merged_main.tex
\begin{table}[t]
  \centering
  \caption{\textbf{Arbitrary-scale super-resolution latency and memory at $\times 4$.} All rows were timed in one session on one NVIDIA H200 GPU as described in Section~\ref{sec:experiments} and Appendix~\ref{app:asr}. The GB-LSR-Scalar-ASR rows are means over three training seeds; the other rows are single checkpoints released by their authors, Thera without its self-ensemble. Peak memory: allocated GPU memory on Urban100 (MiB), weights included; not available for Thera. Speed: per test set, the latency of GB-LSR-Scalar-ASR (base) divided by the latency of the row, then the geometric mean over the three test sets. Parameter counts are in Table~\ref{tab:asr_quality}.}
  \label{tab:asr_latency}
  \footnotesize
  \setlength{\tabcolsep}{4pt}
  \begin{tabular}{lrrrrr}
    \toprule
    Method & \multicolumn{3}{c}{Latency at $\times 4$ (ms) $\downarrow$} & Peak memory & Speed \\
    \cmidrule(lr){2-4}
     & Set14 & B100 & Urban100 & Urban100 (MiB) & vs base $\uparrow$ \\
    \midrule
    GB-LSR-Scalar-ASR (base) & 34.8 & 25.6 & 109.4 & 899 & 1.00$\times$ \\
    GB-LSR-Scalar-ASR-noLE & 17.7 & 14.6 & 52.5 & 900 & 1.93$\times$ \\
    GB-LSR-Scalar-ASR-nf96+noLE & 20.1 & 16.0 & 55.7 & 1{,}109 & 1.76$\times$ \\
    \midrule
    LIIF-RDN & 43.5 & 31.7 & 137.5 & 901 & 0.801$\times$ \\
    LTE-RDN & 42.2 & 31.1 & 130.5 & 911 & 0.829$\times$ \\
    SRNO-RDN & 34.7 & 26.6 & 108.2 & 13{,}692 & 0.991$\times$ \\
    OPE-SR-RDN & 39.0 & 29.4 & 118.8 & 2{,}333 & 0.895$\times$ \\
    \midrule
    LTE-SwinIR & 142.6 & 101.2 & 448.4 & 922 & 0.247$\times$ \\
    LMF & 13.0 & 8.7 & 39.5 & 590 & 2.79$\times$ \\
    SRNO-EDSR & 25.2 & 17.4 & 84.3 & 13{,}613 & 1.38$\times$ \\
    OPE-SR-EDSR & 25.7 & 16.8 & 82.1 & 316 & 1.40$\times$ \\
    GSASR & 88.2 & 53.9 & 427.1 & 8{,}077 & 0.363$\times$ \\
    Thera & 25.6 & 18.3 & 72.5 & N/A & 1.42$\times$ \\
    \bottomrule
  \end{tabular}
\end{table}

%% file: sections/07_limitations.tex
\section{Limitations}
\label{sec:limitations}

Each limitation below points to the evidence behind it.

\paragraph{Perceptual quality is the weakest axis.}
\label{sec:lim:perceptual}

Under one shared LPIPS-AlexNet evaluation, the three GB-LSR rows have
the highest mean LPIPS at $\times 4$ of all methods compared
(Section~\ref{sec:results:asr}). From $\times 2$ to $\times 8$, the base
model has a higher LPIPS than every baseline in $11$ of the $25$ cells, and
LMF, SRNO-EDSR, or OPE-SR-EDSR score worse than it in the others, as
LIIF-RDN also does on Set14 $\times 2$ (Table~\ref{tab:asr_grid_lpips}). The methods that lead on this axis are also
trained with pixel losses, so the gap does not come from generative
synthesis, and we have not identified its cause.

\paragraph{Continuity comes only from the local ensemble.}
\label{sec:lim:boundaries}

The native models and the super-resolution variants trained without the
ensemble are piecewise smooth, with measurable jumps or seams at cell and
patch boundaries, and no model imposes derivative continuity
(Section~\ref{sec:method:boundaries}, Appendix~\ref{app:boundary}).

\paragraph{Per-patch locality is empirically unsupported.}
\label{sec:lim:locality}

Per-patch bandwidth gave no clear gain over one global scalar in the design
study, and after $1{,}000{,}000$ steps the learned
field is near-constant within each image: its median within-image
coefficient of variation is at most $3.7 \times 10^{-4}$ as a three-seed
mean, although $s_e$
may range over $[0.25, 2.0]$, and it passes none of four locality tests
(Appendices~\ref{app:locality}--\ref{app:adaptive_bw_ablation},
Table~\ref{tab:locality},
Figures~\ref{fig:cov_hist}--\ref{fig:logspace_loctests}). The tests
cover $256{\times}256$ native reconstruction only; no per-patch variant
was trained for super-resolution, and we claim no per-patch locality mechanism.

\paragraph{What we do not claim.}
\label{sec:lim:notclaimed}

We do not claim state of the art over the super-resolution literature,
higher reconstruction accuracy than the baselines compared here,
a gain from learning the bandwidth over a well-chosen fixed value,
spatially local bandwidth adaptation, or applicability outside natural
images (Appendices~\ref{app:design} and~\ref{app:asr}).

%% file: sections/08_conclusion.tex
\section{Conclusion}
\label{sec:conclusion}

GB-LSR learns one scalar bandwidth for a fixed-grid local spectral basis,
where Thera learns a global bank of frequencies. Like OPE-SR, it
decodes a coordinate with a fixed-size contraction whose cost does not
grow with image size or output resolution. In super-resolution the
learned bandwidth stays near its start, so the start value matters.

On arbitrary-scale super-resolution, GB-LSR-Scalar-ASR gives up
$0.07$ to $0.79$\,dB of in-distribution PSNR-Y to the three
encoder-matched baselines for speed against LIIF-RDN and LTE-RDN and for
peak memory against SRNO-RDN's released code. Without the local ensemble
it runs $2.41\times$ faster than LIIF-RDN at $\times 4$ and $3.00\times$ at
$\times 8$ with PSNR-Y unchanged within seed variation, at the cost of value
jumps at cell boundaries of $0.22$ and $0.31$ gray levels on average at
$\times 4$ and $\times 8$. Both variants are weakest on perceptual quality.

Because each patch's coefficient block is a fixed-size object
independent of spatial resolution, sharing coefficients across time
could carry the construction to video. Whether one global bandwidth
survives that extension is left for future work.

%% file: sections/appendix.tex
\section{Technical appendices and supplemental material}
\label{app:technical}

This appendix collects material supporting the main paper: the design
study of the representation (Appendix~\ref{app:design}), including the
patch-size study, the locality tests of the per-patch bandwidth, and the
comparison of bandwidth handling
(Appendices~\ref{app:patchsize}--\ref{app:adaptive_bw_ablation});
additional qualitative panels (Appendix~\ref{app:qualitative});
full arbitrary-scale super-resolution details: the timing protocol, the
table of \mbox{GB-LSR-Scalar-ASR} variants, the per-scale grids, the $\times 12$ to
$\times 30$ table, and the cutoff and bandwidth-initialization table
(Appendix~\ref{app:asr}); continuity at cell and patch
boundaries (Appendix~\ref{app:boundary}); reproduction of the released
baselines (Appendix~\ref{app:additional_evidence}); compute resources
(Appendix~\ref{app:compute}); and licenses for existing assets
(Appendix~\ref{app:licenses}).

\subsection{Design study of the representation}
\label{app:design}

This section compares GB-LSR variants against one another under an
otherwise fixed recipe. It is not a comparison against other methods,
and its numbers are not comparable to those of
Section~\ref{sec:results}: the task, the datasets, and the metric
convention all differ (Section~\ref{sec:experiments}).

\paragraph{Training and evaluation.} Every configuration is trained on
$256{\times}256$ crops from a mixture of DTD and DIV2K for
$1{,}000{,}000$ steps at batch size~$8$ with a mean squared error loss and
AdamW
($\beta = (0.9, 0.95)$, no weight decay), a constant learning rate of
$2 \times 10^{-4}$, and gradient-norm clipping at $1.0$, over three
seeds: $54$ runs of $18$ configurations, of which eight hold $s$ at a set
value or learn it from $1.125$ (Table~\ref{tab:bw_sweep}) and three form
the matched-latent control (Table~\ref{tab:native_patchsize}). Quality is
measured on
held-out Kodak, Set14, and Urban100 at $256{\times}256$ and reported as
three-seed means, with sample standard deviations over the seeds where
shown. All of these are
natural images, so the cross-dataset numbers measure a shift within that
regime; medical, satellite, or rendered imagery is not tested. The bandwidth
statistics of Appendix~\ref{app:locality} use the $256$ validation
images held out from the training mixture, the only images for which
the per-patch bandwidth field is stored.

\paragraph{Variants.} All variants share the encoder, partition, and
basis of Section~\ref{sec:method} and differ in how they set the
bandwidth and, for GB-LSR-Full, the cutoff order. GB-LSR-Fixed holds $s$
at the value given in Section~\ref{sec:method:variants} and trains no
spectral parameter. GB-LSR-Scalar learns one global value
$s = \exp\bigl(\log 0.25 + \log 8\,\sigma(r)\bigr) \in [0.25, 2.0]$ from
a raw parameter $r$ that starts at zero, which gives the start value of
Section~\ref{sec:method:variants}. GB-LSR-Bandwidth predicts a
per-patch $s_e$ through the same map from the output $r_e$ of a
zero-initialized linear head on the patch feature, so every $s_e$
starts at the same value. GB-LSR-Full uses the same head to predict, in
addition, a per-patch cutoff order $p_\text{soft} \in [1, 16]$
in the per-mode cutoff of Section~\ref{sec:method:basis}; the other
variants hold $p_\text{soft}$ at $16$. The locality result of
Appendix~\ref{app:locality} holds for this head, a single linear
projection of the patch feature; other head designs are not tested. The
encoder has an input layer, $\log_2 P$ stride-2 downsampling blocks, and an
output layer, all of width $d_\text{feat}$. These four variants use $P = 32$,
$p_{\max} = 16$, and $d_\text{feat} = 128$; the patch-size study varies
these for GB-LSR-Scalar. All values of $s$ in the design study are given
as the implementation uses them. The implementation places the $P$ pixel
centers of a
patch at evenly spaced coordinates from $-1$ to $1$, a normalization by
$(P-1)/2$ rather than the $P/2$ of Section~\ref{sec:method:basis}, so
the bandwidth of Section~\ref{sec:method:basis} is the implementation's
value times $P/(P-1)$ ($32/31$ at $P = 32$, $16/15$ at $P = 16$).

\subsubsection{Patch-size scaling}
\label{app:patchsize}

Holding the spectral cutoff at $p_{\max}=16$ and the feature width at
$d_{\text{feat}}=128$, we compare the default patch size $P = 32$ with
$P = 16$ (Table~\ref{tab:native_patchsize}).
Halving the patch size improves whole-image PSNR-RGB by $6.13$, $5.66$,
and $5.55$\,dB on Kodak, Set14, and Urban100 while using $14.9\%$ fewer
parameters, because the encoder has one stride-2 block fewer. It also
quadruples the per-image latent, from an $8{\times}8$ grid of
$128$-dimensional features ($8{,}192$ values) to a $16{\times}16$ grid
($32{,}768$ values), and the coefficient count from $49{,}152$ to
$196{,}608$, so this comparison does not separate the finer partition
from the larger latent. At $d_\text{feat} = 128$, latency is $1.367$\,ms
at $P = 32$ and $1.358$\,ms at $P = 16$, and peak allocated memory is
$117.8$ and $117.3$\,MiB. A matched-latent control holds the latent at
$16{,}384$ values and changes the feature width with the patch size:
$P = 16$ with $d_\text{feat} = 64$ minus $P = 32$ with
$d_\text{feat} = 256$ is $+0.06$, $-0.13$, and $+0.13$\,dB of whole-image
PSNR-RGB on Kodak, Set14, and Urban100, inside the $0.5$\,dB margin
fixed before the results were known, so the two are comparable in
PSNR-RGB; the $P = 32$ model has $15.8$ times the parameters, and the
$P = 16$ model has the lower LPIPS on all three sets. This pair also
changes the encoder width, the parameter count, and the coefficient
count, so it does not isolate the partition; the gain from halving $P$
at $d_\text{feat} = 128$ comes with the fourfold latent. A third model at
the same latent, $P = 32$ with $p_{\max} = 32$ and $d_\text{feat} = 256$,
also matches the coefficient count of the $P = 16$ model; it is $0.17$
to $0.20$\,dB above that model in PSNR-RGB, inside the same margin, and
has the higher LPIPS.

Absolute PSNR-RGB follows the size of this latent: on Kodak it is
$26.32$, $32.46$, and $36.21$\,dB at $8{,}192$, $32{,}768$, and
$49{,}152$ values, the last for $d_\text{feat} = 192$ at $P = 16$. Each
image is reconstructed from its own latent, so each configuration fixes
the size of that bottleneck; the study compares configurations and does
not benchmark compression or super-resolution. Two further configurations map the trade-off at
$P = 16$. Lowering the cutoff to $p_{\max}=12$ gives a smaller and
faster model at a cost of $0.15$--$0.28$\,dB, and widening the encoder
to $d_{\text{feat}}=192$, which enlarges the latent to $49{,}152$
values, adds $3.75$, $2.95$, and $3.39$\,dB at higher latency and
memory.

\input{tables/native_patchsize.tex}

\subsubsection{Locality of the learned per-patch bandwidth}
\label{app:locality}

GB-LSR-Bandwidth and GB-LSR-Full may set a different bandwidth $s_e$ in
each of the $64$ patches of an image, anywhere in $[0.25, 2.0]$. Four
tests ask whether they use this freedom. Each is computed per seed on
the $256$ validation images and averaged over the three seeds; patches
are labeled smooth, edge, texture, or mixed by their high-frequency
energy and edge density.

\begin{itemize}
  \item T1: the median over images of the within-image coefficient of
    variation (CoV) of $s_e$, its standard deviation over the patches of
    an image divided by their mean, is at least $0.05$.
  \item T2: fewer than half of the images have a within-image range of
    $s_e$ below $5\%$ of the allowed range of $1.75$.
  \item T3: the mean $s_e$ of texture patches differs from that of
    smooth patches by at least $5\%$ of the mean $s_e$.
  \item T4: the within-image share of the variance of $s_e$,
    $\text{frac}_{\text{within}} = \text{var}_{\text{within}} /
    \text{var}_{\text{total}}$, is at least $0.25$, and
    $\text{var}_{\text{within}}$ is at least $10^{-3}$, so that
    noise-level variation cannot pass.
\end{itemize}

\begin{figure}[t]
  \centering
  \begin{minipage}{0.48\linewidth}
    \centering
    \includegraphics[width=\linewidth]{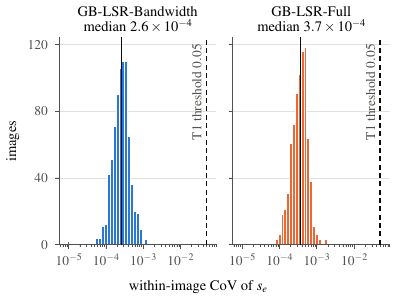}
    \caption{\textbf{Within-image coefficient of variation of the
    per-patch bandwidth.} Per-image CoV of $s_e$ over the $256$
    validation images of each of three seeds
    for GB-LSR-Bandwidth (left, blue) and GB-LSR-Full (right, orange),
    on a logarithmic axis. The solid line marks the median,
    $2.6 \times 10^{-4}$ and $3.7 \times 10^{-4}$ (the mean of the
    per-seed medians), and the dashed line the T1 threshold $0.05$.}
    \label{fig:cov_hist}
  \end{minipage}\hfill
  \begin{minipage}{0.48\linewidth}
    \centering
    \includegraphics[width=\linewidth]{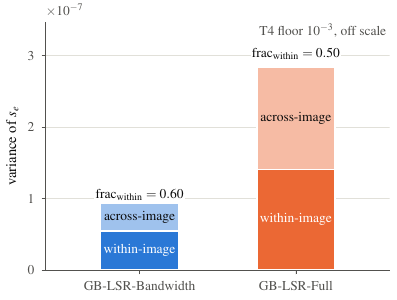}
    \caption{\textbf{Variance decomposition of the per-patch
    bandwidth.} Within-image (darker, lower segment) and across-image
    (lighter, upper segment) variance of $s_e$ for GB-LSR-Bandwidth
    (blue) and GB-LSR-Full (orange), seed means over the $256$
    validation images. Totals are ${9.3 \times 10^{-8}}$ and
    ${2.8 \times 10^{-7}}$; $\text{frac}_{\text{within}}$ (above each
    bar, the mean of the per-seed ratios) is $0.60$ and $0.50$. The T4
    floor of $10^{-3}$ lies off the scale.}
    \label{fig:variance_decomp}
  \end{minipage}
\end{figure}

Neither variant passes any test, in any seed
(Figure~\ref{fig:logspace_loctests}). Each trained model settles on one
value for every patch of every image: averaged over the three seeds,
$s_e$ has mean $0.7180$ and standard deviation $3.0 \times 10^{-4}$
within a model for GB-LSR-Bandwidth, and mean $0.8613$ and standard
deviation $5.3 \times 10^{-4}$ for GB-LSR-Full, whose
adapted cutoff order is just as flat (mean $11.67$, median within-image
CoV at most $3.0 \times 10^{-4}$ in every seed). The median within-image
CoV of $s_e$ is $2.6 \times 10^{-4}$ and $3.7 \times 10^{-4}$
(Figure~\ref{fig:cov_hist}), $191$ and $137$ times below the T1
threshold; every image has a within-image range of $s_e$ below $5\%$ of
the allowed range (T2); and the
texture-smooth gap is $1.3 \times 10^{-4}$ and $4.2 \times 10^{-4}$ of
the mean. For T4, the within-image share of $0.60$ and $0.50$ meets the
$0.25$ threshold, but $\text{var}_{\text{within}}$ is $5.5 \times 10^{-8}$ and
$1.4 \times 10^{-7}$, $18{,}083$ and $7{,}101$ times below the floor
(Figure~\ref{fig:variance_decomp}). Variation across images is as small
($3.8 \times 10^{-8}$ and $1.4 \times 10^{-7}$), so the field acts as a
near-global constant rather than as a per-image or per-patch map.

\subsubsection{Comparison of bandwidth handling}
\label{app:adaptive_bw_ablation}

This comparison uses the held-out sets: whole-image PSNR-RGB, SSIM, and
LPIPS, and PSNR-RGB within the edge, texture, and mixed patches of
Appendix~\ref{app:locality}.

\begin{figure}[t]
  \centering
  \begin{minipage}{0.48\linewidth}
    \centering
    \includegraphics[width=\linewidth]{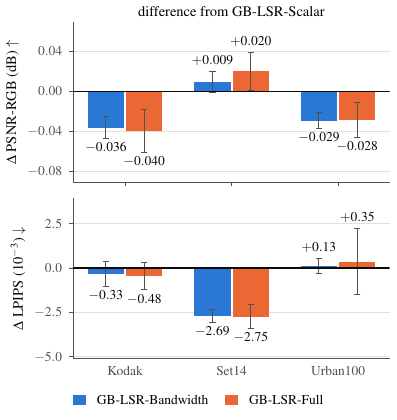}
    \caption{\textbf{Whole-image quality of per-patch bandwidth relative
    to one global scalar.} Held-out PSNR-RGB (top, higher is better) and
    LPIPS in units of $10^{-3}$ (bottom, lower is better) of
    GB-LSR-Bandwidth (blue) and GB-LSR-Full (orange) minus GB-LSR-Scalar
    on Kodak, Set14, and Urban100. Bars are differences of three-seed
    means, labeled with their values; error bars are the sample standard
    deviation of the three seed-paired differences. PSNR-RGB changes
    range from $-0.040$ to $+0.020$\,dB, and LPIPS changes range from
    ${-2.75 \times 10^{-3}}$ to ${+0.35 \times 10^{-3}}$.}
    \label{fig:logspace_metrics}
  \end{minipage}\hfill
  \begin{minipage}{0.48\linewidth}
    \centering
    \includegraphics[width=\linewidth]{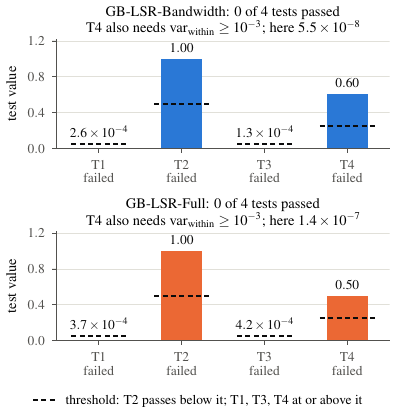}
    \caption{\textbf{Locality-test values of the per-patch bandwidth.}
    T1--T4 for GB-LSR-Bandwidth (top, blue) and GB-LSR-Full (bottom,
    orange), seed means over the $256$ validation images, with each
    threshold as a dashed line (T2 passes below it; T1, T3, and T4 at or
    above it) and the outcome under each label. T1 and T3 are too small
    to show as bars: $2.6 \times 10^{-4}$ and $1.3 \times 10^{-4}$ at
    top, $3.7 \times 10^{-4}$ and $4.2 \times 10^{-4}$ at bottom. T4
    clears its $0.25$ ratio but also needs
    $\text{var}_{\text{within}} \geq 10^{-3}$, against the
    $5.5 \times 10^{-8}$ and $1.4 \times 10^{-7}$ given in the panel
    titles. Both variants pass $0$ of $4$ tests.}
    \label{fig:logspace_loctests}
  \end{minipage}
\end{figure}

A per-patch variant would count as an improvement over GB-LSR-Scalar if
it gained at least $0.30$\,dB of PSNR-RGB on texture or on mixed
patches, or lowered LPIPS by at least $0.015$, while losing at most
$0.20$\,dB on edge patches; these margins were set before the models
were trained. Neither variant meets them on any dataset
(Table~\ref{tab:locality}). The largest texture or mixed gain is
$+0.022$\,dB (GB-LSR-Full, Set14 texture), the largest LPIPS reduction
is $2.75 \times 10^{-3}$ (GB-LSR-Full, Set14), and whole-image PSNR-RGB
changes by $-0.040$ to $+0.020$\,dB
(Figure~\ref{fig:logspace_metrics}). By these margins, per-patch
bandwidth gave no gain over one global scalar in the design study.

GB-LSR-Fixed reaches $22.207$, $19.803$, and $18.888$\,dB on Kodak,
Set14, and Urban100, against $26.325$, $25.350$, and $22.556$\,dB for
GB-LSR-Scalar (standard deviations at most $0.009$\,dB), a gap of
$4.12$, $5.55$, and $3.67$\,dB. GB-LSR-Fixed holds $s = 1.125$, far from the $0.756$
that GB-LSR-Scalar learns from its start at $0.707$. To separate the
value of $s$ from its training, we also trained the configuration of
GB-LSR-Scalar with $s$ held at other values or learned from $1.125$,
three seeds each (Table~\ref{tab:bw_sweep}), and compared runs with the
same seed, which start from the same weights apart from the bandwidth
and see the data in the same order, under a rule fixed before the
results were known: two models match when the mean of their seed-paired
PSNR-RGB differences is at most twice the standard deviation of those
differences in magnitude and every seed whose difference exceeds
$0.01$\,dB in magnitude has the same sign. The seed standard deviations
of these paired differences are $0.002$ to $0.015$\,dB. Held at the learned $0.756$, $s$ matches GB-LSR-Scalar on
Kodak and Set14 and is $0.005$\,dB higher on Urban100. Held at the start
value $0.707$, it matches on Set14 and trails by $0.043$ and $0.033$\,dB
on Kodak and Urban100. Learned from $1.125$ instead, $s$ ends at
$0.812$ after $1{,}000{,}000$ steps, and PSNR-RGB rises above GB-LSR-Fixed
by $4.11$, $5.54$, and $3.68$\,dB, to within $0.009$\,dB of
GB-LSR-Scalar. Held values from
$0.5$ to $0.9$ stay within $0.17$\,dB of one another on every dataset.
The gap between GB-LSR-Fixed and GB-LSR-Scalar therefore comes from the
fixed value, which lies outside that range, and training $s$ matters
only through the value it reaches. At $P = 16$, $s$ held at the learned
$0.658$ is within $0.007$\,dB of GB-LSR-Scalar in the three-seed mean,
and $s$ held at $1.125$,
where the highest modes exceed the Nyquist limit of the $16$-pixel
patch, loses $7.1$ to $9.1$\,dB.

\input{tables/locality.tex}

\input{tables/bw_sweep_appendix.tex}

\subsection{Additional qualitative panels}
\label{app:qualitative}

\begin{figure}[t]
  \centering
  \includegraphics[width=\linewidth]{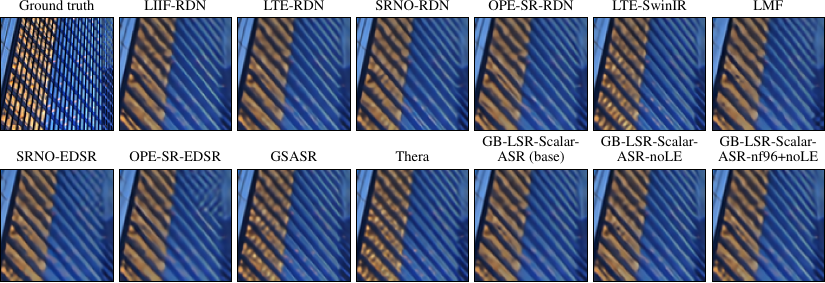}
  \caption{\textbf{All compared methods at $\times 4$.} Urban100
  \texttt{img\_012}, the window of Figure~\ref{fig:qualitative_asr}.
  Baselines use their authors' released checkpoints, Thera without its
  self-ensemble; each GB-LSR tile shows one of its three trained
  models.}
  \label{fig:qualitative}
\end{figure}

\begin{figure}[t]
  \centering
  \includegraphics[width=\linewidth]{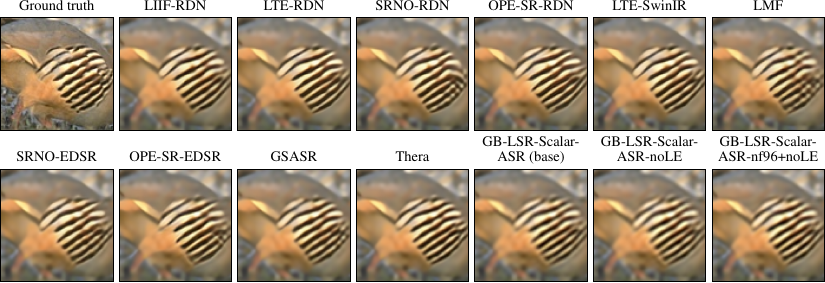}
  \caption{\textbf{All compared methods at $\times 4$, second image.}
  B100 \texttt{8023}, a $96{\times}96$ window chosen by the rule of
  Figure~\ref{fig:qualitative_asr}; tiles as in
  Figure~\ref{fig:qualitative}.}
  \label{fig:qualitative_b100}
\end{figure}

Figure~\ref{fig:qualitative} extends Figure~\ref{fig:qualitative_asr}
to every compared method, and Figure~\ref{fig:qualitative_b100} repeats
it on a B100 image. We picked both images by hand, and each window from
the ground truth alone by the rule of Figure~\ref{fig:qualitative_asr}.

\subsection{Arbitrary-scale super-resolution details}
\label{app:asr}

\paragraph{Methods.} LIIF-RDN, LTE-RDN, SRNO-RDN, LTE-SwinIR, and
OPE-SR-RDN are the checkpoints released by the authors of
LIIF~\citep{chen2021liif}, LTE~\citep{lee2022lte},
SRNO~\citep{wei2023srno}, and OPE-SR~\citep{song2023opesr}, run with
their released model code and inference routine; the five recent methods run
from their released checkpoints and code in the same way. LMF, GSASR, and
Thera each release several EDSR-baseline checkpoints; we use LMF's LM-LIIF
model, GSASR's paper-version checkpoint, and Thera's Pro variant.
Low-resolution generation and metrics are shared, while each baseline's inference
follows its released code: LTE-RDN and LTE-SwinIR clip the cell input to
its $\times 4$ value at larger scales, and LTE-SwinIR pads its input to a
multiple of the window size. GB-LSR-Scalar-ASR is the arbitrary-scale
extension of GB-LSR-Scalar (Section~\ref{sec:method:asr}) on the RDN
encoder~\citep{zhang2018rdn}, with the local spectral decoder
($p_\text{max} = 16$; each coefficient block's support is its own
low-resolution cell without the local ensemble and, with it, the square
two cells wide centered on that cell, so its size in high-resolution
pixels depends on the scale rather than being a fixed patch side) and a
single global trainable scalar bandwidth. The super-resolution bandwidth
uses a softplus parameterization (strictly positive, unbounded above)
initialized at $s = 1.0$, rather than the design study's log-space
sigmoid bound to $[0.25, 2.0]$; from this start the trained value remains
well inside that range ($s \approx 0.88$ on all three seeds;
Table~\ref{tab:asr_sensitivity} gives other starts). As in the released
LIIF training configuration, each training example is a square crop of
side $\mathrm{round}(48k)$ pixels with $k$ drawn uniformly from
$[1, 4]$, downsampled to $48{\times}48$ with PIL bicubic, and the loss is
taken on $2{,}304$ of its pixels drawn without replacement. The released
training code of LIIF, LTE, SRNO, OPE-SR, LMF, and Thera also makes its
low-resolution inputs with PIL bicubic, so our test inputs use the kernel
these models were trained with; GSASR trains with a MATLAB-style bicubic,
which our inputs do not match.

\paragraph{GPU latency protocol.} The latency and memory numbers at
$\times 4$ come from one session on one NVIDIA H200 GPU that runs the
methods one after another, and those at $\times 8$
(Table~\ref{tab:asr_latency_x8}) from a second session of the same kind on
another H200; Section~\ref{sec:experiments} states the statistic. A timed pass
starts from the low-resolution input on the GPU and covers input
normalization, padding, coordinate and cell construction, the query,
cropping, and clamping; low-resolution generation and file input and
output lie outside it. The PyTorch methods run at batch size~1 in single
precision with PyTorch's default TF32 convolutions, without gradients,
automatic mixed precision, compilation, or graph capture, with CUDA
events around each pass after a device synchronization. GB-LSR, the
released LIIF, LTE, and OPE-SR models, and LMF query in chunks of at
most $30{,}000$ points; both SRNO models and GSASR run unchunked, as in
their released code; chunking the queries independently, as the other
methods do, would change the output of SRNO, whose attention averages over
every output position. The released
baselines run their unmodified query code, including its host-side
coordinate construction, and LMF reuses its low-resolution coordinate
grid between calls on images of the same size, as its code does; GB-LSR
also builds its coordinates on the host and recomputes its coefficient
projection for each query chunk, which can only add to its latency. As
in their released code, LTE-SwinIR and GSASR pad their input and Thera
decodes overlapping patches at the image borders, so LTE-SwinIR and GSASR
decode $4$ to $9\%$ more pixels than the output on these test sets at
$\times 4$ and $6$ to $15\%$ and $7$ to $23\%$ more at $\times 8$, respectively, and Thera
$18$ to $75\%$ more at both scales, which can only add to their latency. GSASR runs without its optional
bf16 and tiled modes, on PyTorch 2.5.1. Thera runs on JAX 0.4.11 at its
default precision; its encoder and decoder are XLA-compiled, with
compilation in the warm-up passes, while its patch loop and output
assembly run on the host as in the authors' code, so its timed pass
ends with the output in host memory. It decodes square output patches of
at most $256$ by $256$ pixels and is timed by wall clock with a device
synchronization before and after each pass; JAX reports no
peak allocated memory comparable to PyTorch's, so Thera's memory is not
listed. Memory is the peak allocated GPU memory over the timed passes of
a cell, weights included. To check for drift, one of the three
GB-LSR-Scalar-ASR (base) models was timed again at the end of each session:
in the $\times 4$ session, about seven hours after its first timing, its
latency changed by $-0.16\%$, $-0.19\%$, and $-0.16\%$ on Set14, B100, and
Urban100; in the $\times 8$ session, about three and a half hours later, by
$-0.27\%$, $-0.45\%$, and $-0.09\%$.

\input{tables/asr_family_appendix.tex}

\input{tables/asr_latency_x8_appendix.tex}

\input{tables/asr_grid_psnr_appendix.tex}

\input{tables/asr_grid_ssim_appendix.tex}

\input{tables/asr_grid_lpips_appendix.tex}

\input{tables/asr_ood_ladder_appendix.tex}

\begin{figure}[t]
  \centering
  \includegraphics[width=0.85\linewidth]{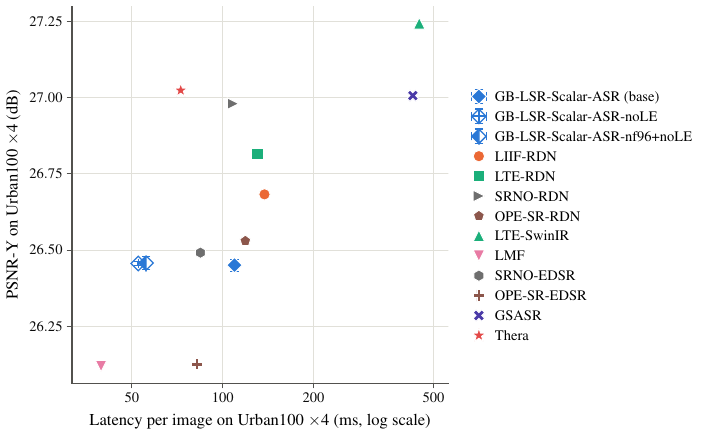}
  \caption{\textbf{PSNR-Y against latency on Urban100 $\times 4$.}
  Latency on a logarithmic axis from the $\times 4$ timing session of
  Section~\ref{sec:experiments}, PSNR-Y from Table~\ref{tab:asr_quality}.
  The GB-LSR points are means over three training seeds, with standard
  deviations of at most $0.04$\,ms and $0.022$\,dB, smaller than the
  markers; every other point is a single released checkpoint, Thera
  without its self-ensemble. GB-LSR-Scalar-ASR-noLE and -nf96+noLE
  overlap near $54$\,ms. Encoders and parameter counts are in
  Table~\ref{tab:asr_quality}.}
  \label{fig:asr_quality_latency_x4}
\end{figure}

\begin{figure}[t]
  \centering
  \includegraphics[width=\linewidth]{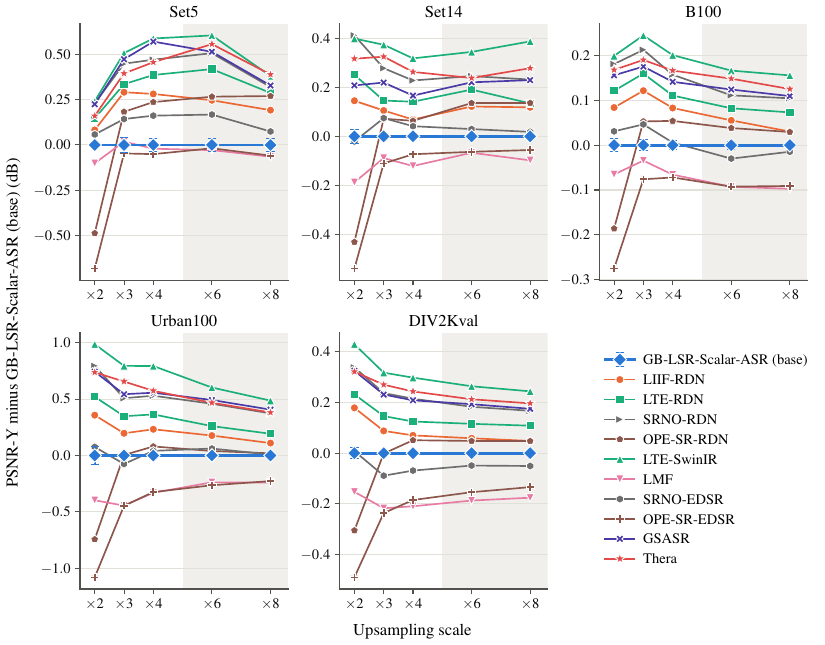}
  \caption{\textbf{PSNR-Y against scale, relative to GB-LSR-Scalar-ASR
  (base).} Each method's PSNR-Y minus the three-seed mean of
  GB-LSR-Scalar-ASR (base) on the same test set and scale, so positive
  values mean a higher PSNR-Y than the base model. Error bars on the base
  model show its seed standard deviation (at most $0.08$\,dB); every other
  line is a single released checkpoint (Thera without its self-ensemble),
  and both OPE-SR models were trained at $\times 4$ only.
  GB-LSR-Scalar-ASR-noLE and -nf96+noLE, within $0.03$ and $0.07$\,dB of
  the base on every cell, are not drawn. Shaded: $\times 6$ and $\times 8$,
  outside the training range of GB-LSR-Scalar-ASR ($\times 1$ to
  $\times 4$). Parameter counts are in Table~\ref{tab:asr_quality}.}
  \label{fig:asr_psnr_vs_scale}
\end{figure}

\paragraph{Encoder width.} Narrowing the RDN encoder to $32$ feature
channels (GB-LSR-Scalar-ASR-nf32) lowers in-distribution PSNR-Y by
$0.068$\,dB on average and by up to $0.153$\,dB, and peak memory from
$899$ to $583$\,MiB; at $48$ channels the loss is $0.023$\,dB, and at
$96$ channels PSNR-Y gains $0.031$\,dB. Speed stays within $5\%$ of the
base model at all three widths (Table~\ref{tab:asr_family}); on a single
test set latency changes by up to $6.5\%$ ($96$ channels, Set14).

\paragraph{Cutoff and bandwidth initialization.} Setting the cutoff to
$p_{\max} = 12$, $20$, or $24$ changes mean in-distribution PSNR-Y by
$-0.006$ to $+0.006$\,dB, and initializing the bandwidth at $s = 0.5$
changes it by $-0.004$\,dB, all within the base model's seed standard
deviation of $0.015$\,dB; initializing it at $s = 2.0$ lowers it by
$2.23$\,dB, and by $3.71$\,dB on Urban100 $\times 3$
(Table~\ref{tab:asr_sensitivity}). The learned bandwidth ends near its
start after $1{,}000{,}000$ steps, at $0.880$ from $1.0$, $0.481$ from
$0.5$, and $1.875$ to $2.053$ over the three seeds started at $2.0$, and the
training loss of the model started at $2.0$ stays above that of the base
model to the end. These results therefore measure sensitivity to the value
of $s$. Unlike the design
study, where $p_{\max} = 12$ costs $0.15$ to $0.28$\,dB
(Appendix~\ref{app:patchsize}), the super-resolution decoder stays
within seed variation at $p_{\max} = 12$.

\input{tables/asr_sensitivity_appendix.tex}

\paragraph{Per-cell speedups.} At $\times 4$ on Set14, B100, and
Urban100, GB-LSR-Scalar-ASR runs faster than LIIF-RDN by $1.25$, $1.24$,
and $1.26\times$ and faster than LTE-SwinIR by $4.10$, $3.95$, and
$4.10\times$; without the ensemble the speedups over LIIF-RDN are
$2.46$, $2.17$, and $2.62\times$. Its speed relative to \mbox{SRNO-RDN} is
$1.00$, $1.04$, and $0.99\times$, and $1.96$, $1.82$, and $2.06\times$
without the ensemble. All ratios are computed
from the seed means behind Table~\ref{tab:asr_latency}.

\paragraph{Scope statement.} This is a quality and efficiency comparison
on the standard arbitrary-scale super-resolution benchmark, over five
test sets, three in-distribution scales, and two out-of-distribution
scales, with four larger ones on the DIV2K validation split for the
models of Table~\ref{tab:asr_ood_ladder}; it is not a survey of the
super-resolution literature. For LIIF, LTE, and SRNO we use the authors'
released checkpoints on the RDN encoder, for LTE also on SwinIR, and for
OPE-SR on a $147$-channel RDN (Section~\ref{sec:experiments}).
Every evaluated recent method releases its code and checkpoints, and
each is evaluated from its EDSR-baseline checkpoint; LMF, GSASR, and
Thera also release RDN checkpoints, and LMF and GSASR SwinIR checkpoints,
which we did not evaluate. Of the recent
methods we screened, seven had code and checkpoints that we located when
we chose the set: the five evaluated here, and CiaoSR and
SSRNO~\citep{han2024ssrno}, which need additional software builds and
were deferred. A-LIIF, CLIT, and HIIF were not part of that screen, and
for LIWT and DIIN we located no repository at the time; these methods
and Meta-SR are not evaluated, and we make no claim about how GB-LSR
compares with them. Our re-implementations of LIIF and LTE appear only in
the reproduction check of Appendix~\ref{app:additional_evidence}. We do
not compare against fixed-scale architectures such as
EDSR~\citep{lim2017edsr}, RCAN~\citep{zhang2018rcan},
NLSN~\citep{mei2021nlsn}, and IPT~\citep{chen2021ipt}, which do not
decode at arbitrary scale, or against per-image neural fields such as
SIREN~\citep{sitzmann2020siren} and WIRE~\citep{saragadam2023wire}, which
are fit to each image rather than predicted by an encoder. Nor is this a claim of higher reconstruction
accuracy: LIIF-RDN, LTE-RDN, SRNO-RDN, LTE-SwinIR, GSASR, and Thera lead GB-LSR-Scalar-ASR on
every in-distribution cell (Section~\ref{sec:results:asr},
Table~\ref{tab:asr_grid_psnr}).

\clearpage
\input{tables/cell_boundary_appendix.tex}

\subsection{Continuity at cell and patch boundaries}
\label{app:boundary}

We measure how the super-resolution output behaves at the boundaries
between low-resolution cells for the three seeds of the base variant and
of the variant trained without the local ensemble, on all $319$ images
of the five benchmarks at $\times 4$ and $\times 8$. Every measure uses
the luma (BT.601 Y) in gray levels out of $255$.

\paragraph{Measures.} The value jump $J(d)$ is the mean absolute
difference between the outputs at distance $d$ on either side of a cell
boundary, with $d$ in units of one cell, over every interior boundary
along both axes. For an output that is continuous in value $J$ shrinks
in proportion to $d$, so $J(10^{-3})/J(10^{-4}) \approx 10$, while a
jump gives a ratio near $1$, and $J(10^{-4})$ is then its size. The
boundary excess $E$ compares the mean absolute difference between
adjacent high-resolution pixels that straddle a cell boundary with the
mean over the two neighboring pixel phases, each first divided by its
ground-truth value, so $E = 1$ means no excess. The band gap is PSNR-Y
over the pixels away from cell boundaries minus PSNR-Y over the pixels
next to one. It is positive for every method, bicubic interpolation
included, because pixels next to a boundary lie farthest from the
low-resolution samples, so only its change between variants is
informative. Bicubic interpolation (cubic convolution with
$a = -0.75$) and nearest-neighbor interpolation of the same input serve
as continuous and discontinuous references, and the base
variant with the ensemble disabled at test time, which it was never
trained for, serves as a positive control. Intervals are $95\%$
bootstrap intervals over seeds and images, paired by image.

\begin{figure}[t]
  \centering
  \includegraphics[width=\linewidth]{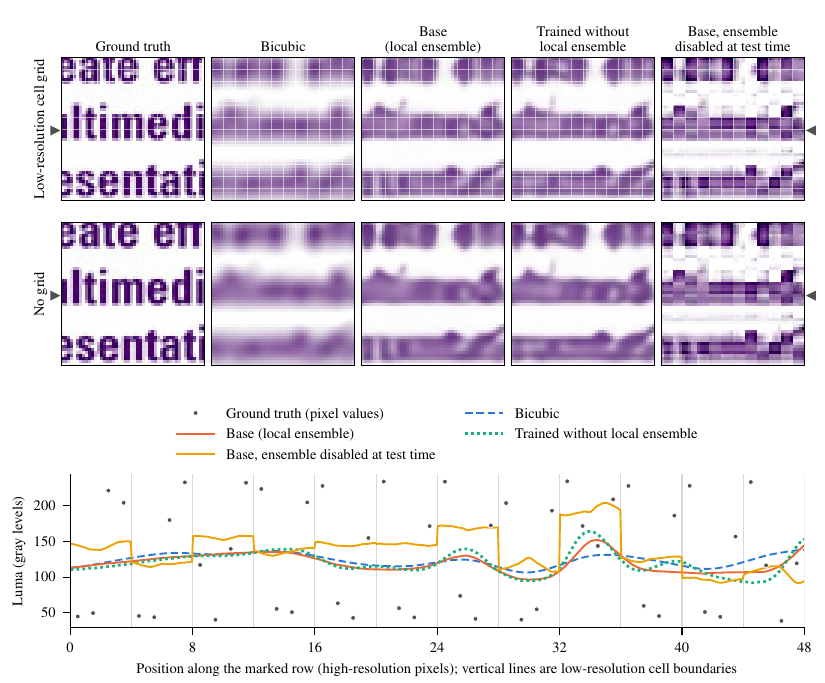}
  \caption{\textbf{Output around low-resolution cell boundaries at
  $\times 4$.} Set14 \texttt{ppt3}, the cell-aligned $48{\times}48$
  window with the largest ground-truth luma gradient energy over Set5 and
  Set14, chosen from the ground truth alone; one trained model of each
  variant (Base: GB-LSR-Scalar-ASR (base); trained without local
  ensemble: GB-LSR-Scalar-ASR-noLE). Top:
  tiles magnified with the low-resolution cell grid; middle: the same
  tiles without the grid; bottom: luma along the row marked by the
  triangles, with cell boundaries as vertical lines. In the tiles the
  base variant and the variant trained without the ensemble are hard to
  tell apart, and neither profile jumps visibly at a cell boundary, while
  the base variant with the ensemble disabled at test time shows what
  large jumps look like.}
  \label{fig:cell_boundary}
\end{figure}

\paragraph{Results.} With the ensemble the output is continuous at
cell boundaries, as for bicubic interpolation: the ratio is $10.0$
(Table~\ref{tab:cell_boundary}). It is also smooth there, since the
same four cells are blended on both sides of a boundary. Without the
ensemble the output jumps by $0.22$ gray levels on average at
$\times 4$ and by $0.31$ at $\times 8$, far less than nearest-neighbor
interpolation or the positive control. $J$ is a mean over boundary
points: per image and trained model it stays below one gray level at
$\times 4$ (at most
$0.97$) and exceeds one on $5$ of the $319$ images at $\times 8$ (at
most $1.26$). At $\times 4$ the jump does not change the boundary excess
or the band gap detectably: both intervals include zero. At $\times 8$,
outside the training range, the boundary excess rises detectably, by
$0.010$, while the band gap narrows, so PSNR-Y next to the boundaries
does not fall relative to the interior. Over the whole image, removing the ensemble changes PSNR-Y
by $+0.002$\,dB [$-0.016$, $+0.019$] over the $15$ in-distribution cells
and by $+0.002$\,dB [$-0.009$, $+0.016$] over the $10$
out-of-distribution cells at $\times 6$ and $\times 8$, with a worst in-distribution cell of
$-0.008$\,dB [$-0.037$, $+0.017$] on Set14 $\times 4$.
Figure~\ref{fig:cell_boundary} shows one window.

The native models decode each patch on its own pixel grid, so we
report only $E$ at their patch boundaries. On the $138$ held-out native
images it is $1.22$ at $P = 16$ and $2.05$ at $P = 32$: relative to the
ground truth, the difference across a patch boundary is $1.22$ and
$2.05$ times that of the neighboring pixel pairs, so the native
reconstructions carry seams at patch boundaries, stronger at $P = 32$.

\subsection{Reproduction of the released baselines}
\label{app:additional_evidence}

To check our
evaluation code, we score the released LIIF-RDN, LTE-RDN, LTE-SwinIR, and
OPE-SR-RDN checkpoints against all $80$ values their authors publish,
from $\times 2$ to $\times 8$, with the low-resolution inputs of the
authors' code: MATLAB-style bicubic at $\times 2$ to $\times 4$, where
our tables use PIL bicubic, and PIL bicubic at $\times 6$ and
$\times 8$. On the original ground truth every score lies within
$0.023$\,dB of the published value; on the ground truth cropped to a
multiple of $12$ pixels that our tables use, within $0.049$\,dB. Our
re-implementations of LIIF-RDN, LTE-RDN, and LTE-SwinIR, trained with
three seeds and
scored like the tables, reproduce LIIF-RDN within $0.07$\,dB of the
released weights on every cell but fall short for LTE: LTE-RDN by
$0.41$\,dB on average in distribution and by up to $1.88$\,dB, and
LTE-SwinIR by $0.33$\,dB and up to $1.71$\,dB. These re-implementations
omit the cell scaling and the bilinear skip connection of the official
LTE decoder, and the paper reports the released checkpoints instead.

\subsection{Compute resources}
\label{app:compute}

Every run used a single GPU on an internal academic cluster. Of the
$54$ design-study runs, $16$ ran on an NVIDIA H100 80\,GB and $38$ on an
NVIDIA H200 SXM 141\,GB, and every arbitrary-scale super-resolution
training run ran on an H200.
Tables~\ref{tab:compute_native} and~\ref{tab:compute_asr} list the
training runs behind the results reported in this paper, with the
wall-clock read from each run's log; evaluation and timing runs are not
included.
\input{tables/compute_total.tex}
Times are single-GPU wall-clock hours, rounded half up to two
decimals; every row, subtotal, and total is summed from unrounded
per-run times and rounded once, so the displayed parts can differ from
their total in the last digit.

\begin{table}[!h]
  \centering
  \caption{\textbf{Design-study training compute.} Training runs behind
  the design-study results, by group and GPU type; each configuration
  trains for $1{,}000{,}000$ steps with three seeds. Per run: the range
  of single-GPU wall-clock hours over the runs of a line; total: their
  sum.}
  \label{tab:compute_native}
  \footnotesize
  \setlength{\tabcolsep}{4pt}
  \begin{tabular}{>{\raggedright\arraybackslash}p{0.52\linewidth}lrrr}
    \toprule
    Training group & GPU & Runs & Per run (h) & Total (h) \\
    \midrule
    GB-LSR variants and patch sizes (Appendix~\ref{app:design}) & H100 & 16 & 11.86--19.16 & 219.13 \\
     & H200 & 5 & 10.49--11.02 & 53.63 \\
    \addlinespace[3pt]
    Held and learned bandwidth (Table~\ref{tab:bw_sweep}) & H200 & 24 & 10.49--10.84 & 254.71 \\
    \addlinespace[3pt]
    Matched-latent control (Table~\ref{tab:native_patchsize}) & H200 & 9 & 10.57--10.90 & 96.09 \\
    \midrule
    Subtotal & H200 & 38 & & 404.43 \\
    Subtotal & H100 & 16 & & 219.13 \\
    Subtotal & all & 54 & & 623.56 \\
    \bottomrule
  \end{tabular}
\end{table}

\begin{table}[!ht]
  \centering
  \caption{\textbf{Arbitrary-scale super-resolution training compute.}
  Training runs behind the super-resolution results, as in
  Table~\ref{tab:compute_native}; each model trains for $1{,}000{,}000$
  steps on DIV2K with three seeds.}
  \label{tab:compute_asr}
  \footnotesize
  \setlength{\tabcolsep}{4pt}
  \begin{tabular}{>{\raggedright\arraybackslash}p{0.52\linewidth}lrrr}
    \toprule
    Training group & GPU & Runs & Per run (h) & Total (h) \\
    \midrule
    GB-LSR-Scalar-ASR (base) & H200 & 3 & 35.93--35.98 & 107.87 \\
    \addlinespace[3pt]
    Re-implemented LIIF-RDN, LTE-RDN (Appendix~\ref{app:additional_evidence}) & H200 & 6 & 36.04--36.20 & 216.74 \\
    \addlinespace[3pt]
    Re-implemented LTE-SwinIR (Appendix~\ref{app:additional_evidence}) & H200 & 3 & 122.21--122.38 & 366.93 \\
    \addlinespace[3pt]
    Encoder-width and local-ensemble variants (Table~\ref{tab:asr_family}) & H200 & 15 & 35.70--36.63 & 540.53 \\
    \addlinespace[3pt]
    32-channel encoder (Table~\ref{tab:asr_family}) & H200 & 3 & 35.92--35.99 & 107.86 \\
    \addlinespace[3pt]
    Cutoff and bandwidth-initialization variants (Table~\ref{tab:asr_sensitivity}) & H200 & 15 & 35.94--36.45 & 543.04 \\
    \midrule
    Subtotal & H200 & 45 & & 1{,}882.98 \\
    \bottomrule
  \end{tabular}
\end{table}

\subsection{Licenses for existing assets}
\label{app:licenses}

Table~\ref{tab:licenses} lists the existing assets used in this paper,
together with their licenses or stated terms of use as retrieved
from the source pages it lists. License names follow SPDX
identifiers where the project ships a formal license file; where no
formal license is published, the project page's stated terms are
summarized instead. All assets are used within their stated terms; no
commercial redistribution is performed.

\begin{table}[t]
  \centering
  \caption{\textbf{Licenses and terms of use for existing assets.}
  Datasets used for training and evaluation, the code and released
  checkpoints of the compared methods (each released with its
  repository under the license shown), and the architectures, metric
  code, and tools used here.}
  \label{tab:licenses}
  \scriptsize
  \setlength{\tabcolsep}{3pt}
  \renewcommand{\arraystretch}{1.2}
  \def\UrlBreaks{\do\/}
  \newcommand{\site}[1]{\href{https://#1}{\nolinkurl{#1}}}
  \begin{tabular}{>{\raggedright\arraybackslash}p{0.265\linewidth}>{\raggedright\arraybackslash}p{0.31\linewidth}>{\raggedright\arraybackslash}p{0.38\linewidth}}
    \toprule
    Asset & License or terms of use & URL \\
    \midrule
    \multicolumn{3}{l}{Datasets} \\
    DIV2K~\citep{agustsson2017div2k} & For academic research purpose only & \site{data.vision.ee.ethz.ch/cvl/DIV2K/} \\
    DTD~\citep{cimpoi2014dtd} & Released for research purposes (no formal license name on project page) & \site{www.robots.ox.ac.uk/~vgg/data/dtd/} \\
    Set5~\citep{bevilacqua2012set5} & No license published; available from the SelfExSR repository of \citet{huang2015urban100} & \site{github.com/jbhuang0604/SelfExSR} \\
    Set14~\citep{zeyde2010set14} & No license published; available from the SelfExSR repository of \citet{huang2015urban100} & \site{github.com/jbhuang0604/SelfExSR} \\
    B100~\citep{martin2001bsds} & Non-commercial research and educational purposes (BSDS300 project page) & \site{www2.eecs.berkeley.edu/Research/Projects/CS/vision/grouping/segbench/} \\
    Urban100~\citep{huang2015urban100} & No license published; available from the SelfExSR repository of \citet{huang2015urban100} & \site{github.com/jbhuang0604/SelfExSR} \\
    Kodak~\citep{kodak} & Released by Eastman Kodak Company for unrestricted usage (per project-page maintainer's statement) & \site{r0k.us/graphics/kodak/} \\
    \midrule
    \multicolumn{3}{l}{Code and released checkpoints of the compared methods} \\
    LIIF~\citep{chen2021liif} & BSD-3-Clause & \site{github.com/yinboc/liif} \\
    LTE~\citep{lee2022lte} & BSD-3-Clause & \site{github.com/jaewon-lee-b/lte} \\
    OPE-SR~\citep{song2023opesr} & BSD-3-Clause & \site{github.com/gaochao-s/ope-sr} \\
    LMF~\citep{he2024lmf} & Apache-2.0 & \site{github.com/HeZongyao/LMF} \\
    SRNO~\citep{wei2023srno} & MIT & \site{github.com/2y7c3/Super-Resolution-Neural-Operator} \\
    GSASR~\citep{chen2025gsasr} & Apache-2.0 & \site{github.com/chrisdud0257/gsasr} \\
    Thera~\citep{becker2024thera} & Apache-2.0 & \site{github.com/prs-eth/thera} \\
    \midrule
    \multicolumn{3}{l}{Architectures, metrics, and tools} \\
    RDN encoder~\citep{zhang2018rdn} & No license file in repository (used solely as architectural reference) & \site{github.com/yulunzhang/RDN} \\
    EDSR-baseline encoder~\citep{lim2017edsr} & MIT & \site{github.com/sanghyun-son/EDSR-PyTorch} \\
    SwinIR encoder~\citep{liang2021swinir} & Apache-2.0 & \site{github.com/JingyunLiang/SwinIR} \\
    LPIPS~\citep{zhang2018lpips} & BSD-2-Clause & \site{github.com/richzhang/PerceptualSimilarity} \\
    scikit-image (SSIM) & BSD-3-Clause & \site{github.com/scikit-image/scikit-image} \\
    MATLAB-style bicubic resize & MIT & \site{github.com/fatheral/matlab_imresize} \\
    \bottomrule
  \end{tabular}
\end{table}

Datasets without an explicit license file are used following the
access terms stated on the pages listed in Table~\ref{tab:licenses}. Code
repositories without an explicit license are used solely as
architectural reference for the comparison set
(Appendix~\ref{app:asr}); no upstream code is redistributed in
this work.

%% file: tables/native_patchsize.tex
\begin{table}[t]
  \centering
  \caption{\textbf{Patch-size scaling in the design study.} GB-LSR-Scalar at patch size $P$, spectral cutoff $p_{\max}$, and feature width $d_{\text{feat}}$. Whole-image PSNR-RGB (dB) on held-out Kodak, Set14, and Urban100 is given as the mean and sample standard deviation over three seeds. Latent is the number of encoder feature values per $256{\times}256$ image, $(256/P)^2 d_{\text{feat}}$; the coefficient count per image, $3(256/P)^2 p_{\max}^2$, is $49{,}152$, $196{,}608$, $110{,}592$, and $196{,}608$ for the first four rows and $49{,}152$, $196{,}608$, and $196{,}608$ for the three rows of the matched-latent control below them. Latency (ms) and peak allocated GPU memory (MiB) are measured in the timing session of Table~\ref{tab:asr_latency} (Section~\ref{sec:experiments}), averaged over the $138$ held-out images and the three seeds; the matched-latent rows were not timed.}
  \label{tab:native_patchsize}
  \scriptsize
  \setlength{\tabcolsep}{3pt}
  \begin{tabular}{rrrrrr@{\,}lr@{\,}lr@{\,}lrr}
    \toprule
     & & & & & \multicolumn{6}{c}{PSNR-RGB (dB) $\uparrow$} & Latency & Memory \\
    \cmidrule(lr){6-11}
    $P$ & $p_{\max}$ & $d_{\text{feat}}$ & Parameters & Latent & \multicolumn{2}{c}{Kodak} & \multicolumn{2}{c}{Set14} & \multicolumn{2}{c}{Urban100} & (ms) $\downarrow$ & (MiB) $\downarrow$ \\
    \midrule
    32 & 16 & 128 & 989{,}955 & 8{,}192 & 26.325 & $\pm$0.005 & 25.350 & $\pm$0.009 & 22.556 & $\pm$0.005 & 1.367 & 117.8 \\
    16 & 16 & 128 & 842{,}115 & 32{,}768 & 32.455 & $\pm$0.021 & 31.013 & $\pm$0.016 & 28.102 & $\pm$0.011 & 1.358 & 117.3 \\
    16 & 12 & 128 & 798{,}771 & 32{,}768 & 32.197 & $\pm$0.026 & 30.862 & $\pm$0.023 & 27.822 & $\pm$0.017 & 1.302 & 117.1 \\
    16 & 16 & 192 & 1{,}815{,}747 & 49{,}152 & 36.209 & $\pm$0.014 & 33.963 & $\pm$0.007 & 31.497 & $\pm$0.002 & 2.193 & 162.4 \\
    \midrule
    32 & 16 & 256 & 3{,}748{,}611 & 16{,}384 & 28.417 & $\pm$0.006 & 27.627 & $\pm$0.006 & 24.435 & $\pm$0.010 & & \\
    16 & 16 & 64 & 237{,}123 & 16{,}384 & 28.479 & $\pm$0.012 & 27.499 & $\pm$0.008 & 24.568 & $\pm$0.004 & & \\
    32 & 32 & 256 & 4{,}340{,}739 & 16{,}384 & 28.675 & $\pm$0.005 & 27.673 & $\pm$0.009 & 24.761 & $\pm$0.005 & & \\
    \bottomrule
  \end{tabular}
\end{table}

%% file: tables/locality.tex
\begin{table}[t]
  \centering
  \caption{\textbf{Per-patch bandwidth against one global scalar in the design study.} Held-out PSNR-RGB (dB) over the whole image and within edge, texture, and mixed patches (averaged over the images that contain them), with whole-image SSIM and LPIPS. The GB-LSR-Scalar rows give three-seed means; the per-patch rows give the change from GB-LSR-Scalar, the difference of three-seed means. A per-patch variant would count as an improvement with a texture or mixed gain of at least $0.30$\,dB or an LPIPS reduction of at least $0.015$, together with an edge loss of at most $0.20$\,dB; neither meets this on any dataset.}
  \label{tab:locality}
  \footnotesize
  \setlength{\tabcolsep}{4pt}
  \begin{tabular}{llrrrrrr}
    \toprule
     & & \multicolumn{4}{c}{PSNR-RGB (dB) $\uparrow$} & & \\
    \cmidrule(lr){3-6}
    Dataset & Variant & Whole image & Edge & Texture & Mixed & SSIM $\uparrow$ & LPIPS $\downarrow$ \\
    \midrule
    Kodak & GB-LSR-Scalar & 26.325 & 25.186 & 23.111 & 29.462 & 0.7401 & 0.3942 \\
     & GB-LSR-Bandwidth & $-0.036$ & $-0.033$ & $-0.028$ & $-0.043$ & $-0.0016$ & $-0.0003$ \\
     & GB-LSR-Full & $-0.040$ & $-0.037$ & $-0.078$ & $-0.044$ & $-0.0025$ & $-0.0005$ \\
    \midrule
    Set14 & GB-LSR-Scalar & 25.350 & 24.447 & 23.825 & 27.969 & 0.7297 & 0.3497 \\
     & GB-LSR-Bandwidth & $+0.009$ & $+0.020$ & $-0.004$ & $-0.008$ & $+0.0003$ & $-0.0027$ \\
     & GB-LSR-Full & $+0.020$ & $+0.027$ & $+0.022$ & $+0.002$ & $+0.0001$ & $-0.0028$ \\
    \midrule
    Urban100 & GB-LSR-Scalar & 22.556 & 21.452 & 21.000 & 25.006 & 0.6721 & 0.4021 \\
     & GB-LSR-Bandwidth & $-0.029$ & $-0.035$ & $-0.030$ & $-0.038$ & $-0.0023$ & $+0.0001$ \\
     & GB-LSR-Full & $-0.028$ & $-0.031$ & $-0.026$ & $-0.031$ & $-0.0030$ & $+0.0003$ \\
    \bottomrule
  \end{tabular}
\end{table}

%% file: tables/bw_sweep_appendix.tex
\begin{table}[t]
  \centering
  \caption{\textbf{Held and learned bandwidth in the design study.} The configuration of GB-LSR-Scalar ($p_{\max} = 16$, $d_{\text{feat}} = 128$) with $s$ held at the listed value or learned from the listed start, three seeds each, $1{,}000{,}000$ steps; values of $s$ are given as the implementation uses them (Appendix~\ref{app:design}). Whole-image PSNR-RGB (dB), SSIM, and LPIPS on the held-out sets are three-seed means; the seed standard deviation of PSNR-RGB is at most 0.021\,dB. At $P = 32$, $s$ held at $1.125$ is \mbox{GB-LSR-Fixed}, and at both patch sizes $s$ learned from $0.707$ is GB-LSR-Scalar.}
  \label{tab:bw_sweep}
  \scriptsize
  \setlength{\tabcolsep}{3pt}
  \begin{tabular}{rlrrrrrrrrr}
    \toprule
    $P$ & Bandwidth $s$ & \multicolumn{3}{c}{Kodak} & \multicolumn{3}{c}{Set14} & \multicolumn{3}{c}{Urban100} \\
    \cmidrule(lr){3-5} \cmidrule(lr){6-8} \cmidrule(lr){9-11}
     &  & PSNR $\uparrow$ & SSIM $\uparrow$ & LPIPS $\downarrow$ & PSNR $\uparrow$ & SSIM $\uparrow$ & LPIPS $\downarrow$ & PSNR $\uparrow$ & SSIM $\uparrow$ & LPIPS $\downarrow$ \\
    \midrule
    32 & held at $0.500$ & 26.193 & 0.7314 & 0.3860 & 25.338 & 0.7262 & 0.3451 & 22.400 & 0.6605 & 0.3931 \\
     & held at $0.650$ & 26.252 & 0.7367 & 0.3947 & 25.352 & 0.7298 & 0.3474 & 22.496 & 0.6684 & 0.4040 \\
     & held at $0.707$ & 26.282 & 0.7379 & 0.3956 & 25.349 & 0.7295 & 0.3488 & 22.523 & 0.6693 & 0.4035 \\
     & held at $0.756$ & 26.323 & 0.7399 & 0.3947 & 25.353 & 0.7299 & 0.3503 & 22.561 & 0.6724 & 0.4024 \\
     & held at $0.900$ & 26.312 & 0.7388 & 0.3915 & 25.316 & 0.7272 & 0.3434 & 22.558 & 0.6712 & 0.4032 \\
     & held at $1.125$ & 22.207 & 0.6355 & 0.4714 & 19.803 & 0.5785 & 0.4699 & 18.888 & 0.5609 & 0.4556 \\
     & learned from $0.707$ to $0.756$ & 26.325 & 0.7401 & 0.3942 & 25.350 & 0.7297 & 0.3497 & 22.556 & 0.6721 & 0.4021 \\
     & learned from $1.125$ to $0.812$ & 26.322 & 0.7403 & 0.3947 & 25.348 & 0.7299 & 0.3496 & 22.564 & 0.6728 & 0.4029 \\
    \midrule
    16 & held at $0.658$ & 32.462 & 0.9273 & 0.0278 & 31.012 & 0.8954 & 0.0290 & 28.101 & 0.8870 & 0.0329 \\
     & held at $1.125$ & 24.661 & 0.7824 & 0.1963 & 21.904 & 0.6911 & 0.2792 & 20.955 & 0.7169 & 0.2243 \\
     & learned from $0.707$ to $0.658$ & 32.455 & 0.9273 & 0.0280 & 31.013 & 0.8955 & 0.0291 & 28.102 & 0.8871 & 0.0330 \\
    \bottomrule
  \end{tabular}
\end{table}

%% file: tables/asr_family_appendix.tex
\begin{table}[t]
  \centering
  \caption{\textbf{Arbitrary-scale super-resolution: GB-LSR-Scalar-ASR variants.} Every row is a mean over three training seeds. noLE: trained and evaluated without the local ensemble; nf$N$: RDN encoder with base feature width $N$ instead of $64$ (growth rate $64$). Params: trainable parameters in millions. Latency (ms) at $\times 4$ and peak allocated GPU memory on Urban100 $\times 4$ (MiB) are measured as described in Section~\ref{sec:experiments}; speed: per test set, the latency of GB-LSR-Scalar-ASR (base) divided by that of the variant, then the geometric mean over the three test sets. $\Delta$PSNR-Y: variant minus base (dB), scored as in Table~\ref{tab:asr_quality}, averaged over the $15$ in-distribution (ID) cells ($\times 2$, $\times 3$, and $\times 4$ on the five test sets) and the $10$ out-of-distribution (OOD) cells ($\times 6$ and $\times 8$); ID worst: the lowest $\Delta$PSNR-Y over the in-distribution cells.}
  \label{tab:asr_family}
  \scriptsize
  \setlength{\tabcolsep}{3pt}
  \begin{tabular}{lrrrrrrrrr}
    \toprule
    Variant & Params & \multicolumn{3}{c}{Latency at $\times 4$ (ms) $\downarrow$} & Peak memory & Speed & \multicolumn{3}{c}{$\Delta$PSNR-Y vs base (dB) $\uparrow$} \\
    \cmidrule(lr){3-5} \cmidrule(lr){8-10}
     & (M) & Set14 & B100 & Urban100 & (MiB) $\downarrow$ & vs base $\uparrow$ & ID mean & OOD mean & ID worst \\
    \midrule
    GB-LSR-Scalar-ASR (base) & 22.02 & 34.8 & 25.6 & 109.4 & 899 & 1.00$\times$ &  &  &  \\
    GB-LSR-Scalar-ASR-noLE & 22.02 & 17.7 & 14.6 & 52.5 & 900 & 1.93$\times$ & $+0.002$ & $+0.002$ & $-0.008$ \\
    GB-LSR-Scalar-ASR-nf32 & 19.22 & 34.8 & 25.1 & 106.9 & 583 & 1.01$\times$ & $-0.068$ & $-0.061$ & $-0.153$ \\
    GB-LSR-Scalar-ASR-nf48 & 20.61 & 35.5 & 25.3 & 109.8 & 730 & 0.996$\times$ & $-0.023$ & $-0.026$ & $-0.067$ \\
    GB-LSR-Scalar-ASR-nf96 & 24.93 & 37.1 & 27.0 & 112.7 & 1{,}109 & 0.953$\times$ & $+0.031$ & $+0.020$ & $+0.010$ \\
    GB-LSR-Scalar-ASR-nf48+noLE & 20.61 & 18.8 & 14.4 & 52.9 & 730 & 1.90$\times$ & $-0.034$ & $-0.026$ & $-0.073$ \\
    GB-LSR-Scalar-ASR-nf96+noLE & 24.93 & 20.1 & 16.0 & 55.7 & 1{,}109 & 1.76$\times$ & $+0.016$ & $+0.017$ & $-0.006$ \\
    \bottomrule
  \end{tabular}
\end{table}

%% file: tables/asr_latency_x8_appendix.tex
\begin{table}[t]
  \centering
  \caption{\textbf{Arbitrary-scale super-resolution latency and memory at $\times 8$.} All rows were timed in a second session on one NVIDIA H200 GPU with the protocol of Table~\ref{tab:asr_latency} (Section~\ref{sec:experiments} and Appendix~\ref{app:asr}), on the same test sets at $\times 8$. The GB-LSR-Scalar-ASR rows are means over three training seeds; the other rows are single checkpoints released by their authors, Thera without its self-ensemble. Peak memory: allocated GPU memory on Urban100 (MiB), weights included; not available for Thera. Speed: per test set, the latency of GB-LSR-Scalar-ASR (base) divided by the latency of the row, then the geometric mean over the three test sets. In this session the seed standard deviation of a GB-LSR-Scalar-ASR latency is at most $5.5\%$ of its mean ($2.1\%$ for the base variant), against at most $0.13\%$ for the same rows at $\times 4$. PSNR-Y at $\times 8$ is in Table~\ref{tab:asr_grid_psnr}.}
  \label{tab:asr_latency_x8}
  \footnotesize
  \setlength{\tabcolsep}{4pt}
  \begin{tabular}{lrrrrr}
    \toprule
    Method & \multicolumn{3}{c}{Latency at $\times 8$ (ms) $\downarrow$} & Peak memory & Speed \\
    \cmidrule(lr){2-4}
     & Set14 & B100 & Urban100 & Urban100 (MiB) & vs base $\uparrow$ \\
    \midrule
    GB-LSR-Scalar-ASR (base) & 30.0 & 21.5 & 86.6 & 408 & 1.00$\times$ \\
    GB-LSR-Scalar-ASR-noLE & 13.2 & 10.8 & 31.0 & 408 & 2.33$\times$ \\
    GB-LSR-Scalar-ASR-nf96+noLE & 14.8 & 12.7 & 34.6 & 409 & 2.05$\times$ \\
    \midrule
    LIIF-RDN & 38.2 & 27.0 & 115.8 & 423 & 0.777$\times$ \\
    LTE-RDN & 37.4 & 26.6 & 111.5 & 433 & 0.795$\times$ \\
    SRNO-RDN & 30.2 & 22.5 & 90.6 & 13{,}676 & 0.969$\times$ \\
    OPE-SR-RDN & 30.1 & 21.5 & 88.4 & 695 & 0.992$\times$ \\
    \midrule
    LTE-SwinIR & 77.4 & 57.1 & 202.1 & 363 & 0.397$\times$ \\
    LMF & 12.9 & 8.7 & 39.3 & 197 & 2.33$\times$ \\
    SRNO-EDSR & 25.0 & 17.3 & 82.8 & 13{,}597 & 1.16$\times$ \\
    OPE-SR-EDSR & 24.1 & 15.8 & 75.3 & 136 & 1.25$\times$ \\
    GSASR & 39.4 & 33.1 & 117.2 & 2{,}180 & 0.715$\times$ \\
    Thera & 22.4 & 17.1 & 61.9 & N/A & 1.33$\times$ \\
    \bottomrule
  \end{tabular}
\end{table}

%% file: tables/asr_grid_psnr_appendix.tex
\begin{table}[p]
  \centering
  \caption{\textbf{Arbitrary-scale super-resolution quality from $\times 2$ to $\times 8$: PSNR-Y (dB), higher is better.} Scored as in Section~\ref{sec:experiments}; DIV2Kval is the DIV2K validation set. The GB-LSR-Scalar-ASR rows give the mean and sample standard deviation over three training seeds; the other rows are single checkpoints released by their authors. OPE-SR-RDN and OPE-SR-EDSR were trained at $\times 4$ only. Thera is scored without its self-ensemble. Bold: best value in each column of each test set. Row names, encoders, and parameter counts follow Table~\ref{tab:asr_quality}.}
  \label{tab:asr_grid_psnr}
  \scriptsize
  \setlength{\tabcolsep}{2.5pt}
  \begin{tabular}{llr@{\,}lr@{\,}lr@{\,}lr@{\,}lr@{\,}l}
    \toprule
    Test set & Method & \multicolumn{2}{c}{$\times 2$} & \multicolumn{2}{c}{$\times 3$} & \multicolumn{2}{c}{$\times 4$} & \multicolumn{2}{c}{$\times 6$} & \multicolumn{2}{c}{$\times 8$} \\
    \midrule
    Set5 & GB-LSR-Scalar-ASR (base) & 38.057 & $\pm$0.011 & 34.373 & $\pm$0.041 & 32.210 & $\pm$0.036 & 28.878 & $\pm$0.036 & 26.932 & $\pm$0.035 \\
     & GB-LSR-Scalar-ASR-noLE & 38.059 & $\pm$0.027 & 34.376 & $\pm$0.058 & 32.212 & $\pm$0.045 & 28.889 & $\pm$0.039 & 26.958 & $\pm$0.033 \\
     & GB-LSR-Scalar-ASR-nf96+noLE & 38.065 & $\pm$0.030 & 34.410 & $\pm$0.021 & 32.213 & $\pm$0.027 & 28.915 & $\pm$0.033 & 26.986 & $\pm$0.029 \\
     & LIIF-RDN & 38.140 &  & 34.664 &  & 32.491 &  & 29.125 &  & 27.124 &  \\
     & LTE-RDN & 38.204 &  & 34.709 &  & 32.596 &  & 29.297 &  & 27.219 &  \\
     & SRNO-RDN & 38.287 &  & 34.822 &  & 32.681 &  & 29.385 &  & 27.252 &  \\
     & OPE-SR-RDN & 37.570 &  & 34.556 &  & 32.446 &  & 29.145 &  & 27.202 &  \\
     & LTE-SwinIR & \textbf{38.299} &  & \textbf{34.880} &  & \textbf{32.798} &  & \textbf{29.483} &  & 27.313 &  \\
     & LMF & 37.956 &  & 34.391 &  & 32.189 &  & 28.850 &  & 26.867 &  \\
     & SRNO-EDSR & 38.115 &  & 34.516 &  & 32.372 &  & 29.046 &  & 27.006 &  \\
     & OPE-SR-EDSR & 37.375 &  & 34.326 &  & 32.159 &  & 28.860 &  & 26.873 &  \\
     & GSASR & 38.282 &  & 34.846 &  & 32.780 &  & 29.393 &  & 27.260 &  \\
     & Thera & 38.217 &  & 34.768 &  & 32.667 &  & 29.435 &  & \textbf{27.321} &  \\
    \midrule
    Set14 & GB-LSR-Scalar-ASR (base) & 33.833 & $\pm$0.029 & 30.421 & $\pm$0.011 & 28.734 & $\pm$0.013 & 26.509 & $\pm$0.011 & 25.018 & $\pm$0.012 \\
     & GB-LSR-Scalar-ASR-noLE & 33.830 & $\pm$0.046 & 30.431 & $\pm$0.019 & 28.726 & $\pm$0.012 & 26.506 & $\pm$0.021 & 25.015 & $\pm$0.014 \\
     & GB-LSR-Scalar-ASR-nf96+noLE & 33.849 & $\pm$0.039 & 30.449 & $\pm$0.023 & 28.750 & $\pm$0.007 & 26.508 & $\pm$0.009 & 25.018 & $\pm$0.013 \\
     & LIIF-RDN & 33.979 &  & 30.526 &  & 28.803 &  & 26.631 &  & 25.136 &  \\
     & LTE-RDN & 34.085 &  & 30.567 &  & 28.875 &  & 26.700 &  & 25.152 &  \\
     & SRNO-RDN & \textbf{34.246} &  & 30.699 &  & 28.962 &  & 26.755 &  & 25.249 &  \\
     & OPE-SR-RDN & 33.402 &  & 30.493 &  & 28.797 &  & 26.645 &  & 25.154 &  \\
     & LTE-SwinIR & 34.233 &  & \textbf{30.795} &  & \textbf{29.052} &  & \textbf{26.854} &  & \textbf{25.405} &  \\
     & LMF & 33.646 &  & 30.332 &  & 28.613 &  & 26.442 &  & 24.920 &  \\
     & SRNO-EDSR & 33.813 &  & 30.495 &  & 28.775 &  & 26.539 &  & 25.036 &  \\
     & OPE-SR-EDSR & 33.294 &  & 30.310 &  & 28.661 &  & 26.446 &  & 24.962 &  \\
     & GSASR & 34.041 &  & 30.640 &  & 28.900 &  & 26.730 &  & 25.247 &  \\
     & Thera & 34.150 &  & 30.746 &  & 28.997 &  & 26.747 &  & 25.296 &  \\
    \midrule
    B100 & GB-LSR-Scalar-ASR (base) & 32.254 & $\pm$0.015 & 29.155 & $\pm$0.012 & 27.671 & $\pm$0.010 & 25.925 & $\pm$0.009 & 24.874 & $\pm$0.006 \\
     & GB-LSR-Scalar-ASR-noLE & 32.254 & $\pm$0.006 & 29.154 & $\pm$0.001 & 27.674 & $\pm$0.007 & 25.927 & $\pm$0.007 & 24.873 & $\pm$0.004 \\
     & GB-LSR-Scalar-ASR-nf96+noLE & 32.258 & $\pm$0.011 & 29.149 & $\pm$0.004 & 27.674 & $\pm$0.009 & 25.930 & $\pm$0.011 & 24.886 & $\pm$0.007 \\
     & LIIF-RDN & 32.338 &  & 29.277 &  & 27.754 &  & 25.980 &  & 24.904 &  \\
     & LTE-RDN & 32.375 &  & 29.315 &  & 27.782 &  & 26.007 &  & 24.947 &  \\
     & SRNO-RDN & 32.435 &  & 29.368 &  & 27.829 &  & 26.036 &  & 24.978 &  \\
     & OPE-SR-RDN & 32.068 &  & 29.208 &  & 27.726 &  & 25.963 &  & 24.903 &  \\
     & LTE-SwinIR & \textbf{32.453} &  & \textbf{29.400} &  & \textbf{27.872} &  & \textbf{26.091} &  & \textbf{25.030} &  \\
     & LMF & 32.188 &  & 29.121 &  & 27.606 &  & 25.832 &  & 24.777 &  \\
     & SRNO-EDSR & 32.285 &  & 29.201 &  & 27.677 &  & 25.895 &  & 24.859 &  \\
     & OPE-SR-EDSR & 31.978 &  & 29.079 &  & 27.599 &  & 25.833 &  & 24.783 &  \\
     & GSASR & 32.410 &  & 29.330 &  & 27.813 &  & 26.049 &  & 24.984 &  \\
     & Thera & 32.422 &  & 29.345 &  & 27.838 &  & 26.073 &  & 24.999 &  \\
    \midrule
    Urban100 & GB-LSR-Scalar-ASR (base) & 32.529 & $\pm$0.078 & 28.630 & $\pm$0.025 & 26.450 & $\pm$0.020 & 24.021 & $\pm$0.005 & 22.673 & $\pm$0.011 \\
     & GB-LSR-Scalar-ASR-noLE & 32.535 & $\pm$0.029 & 28.642 & $\pm$0.007 & 26.456 & $\pm$0.007 & 24.023 & $\pm$0.003 & 22.665 & $\pm$0.009 \\
     & GB-LSR-Scalar-ASR-nf96+noLE & 32.590 & $\pm$0.016 & 28.639 & $\pm$0.018 & 26.457 & $\pm$0.022 & 24.035 & $\pm$0.007 & 22.692 & $\pm$0.002 \\
     & LIIF-RDN & 32.886 &  & 28.826 &  & 26.682 &  & 24.198 &  & 22.783 &  \\
     & LTE-RDN & 33.049 &  & 28.977 &  & 26.814 &  & 24.282 &  & 22.865 &  \\
     & SRNO-RDN & 33.322 &  & 29.137 &  & 26.980 &  & 24.480 &  & 23.046 &  \\
     & OPE-SR-RDN & 31.786 &  & 28.634 &  & 26.530 &  & 24.061 &  & 22.692 &  \\
     & LTE-SwinIR & \textbf{33.514} &  & \textbf{29.424} &  & \textbf{27.243} &  & \textbf{24.623} &  & \textbf{23.159} &  \\
     & LMF & 32.132 &  & 28.185 &  & 26.120 &  & 23.783 &  & 22.433 &  \\
     & SRNO-EDSR & 32.604 &  & 28.555 &  & 26.491 &  & 24.083 &  & 22.681 &  \\
     & OPE-SR-EDSR & 31.449 &  & 28.184 &  & 26.126 &  & 23.757 &  & 22.446 &  \\
     & GSASR & 33.271 &  & 29.173 &  & 27.006 &  & 24.513 &  & 23.079 &  \\
     & Thera & 33.265 &  & 29.285 &  & 27.023 &  & 24.488 &  & 23.054 &  \\
    \midrule
    DIV2Kval & GB-LSR-Scalar-ASR (base) & 36.325 & $\pm$0.022 & 32.661 & $\pm$0.005 & 30.683 & $\pm$0.010 & 28.408 & $\pm$0.004 & 27.048 & $\pm$0.005 \\
     & GB-LSR-Scalar-ASR-noLE & 36.327 & $\pm$0.012 & 32.667 & $\pm$0.003 & 30.681 & $\pm$0.004 & 28.405 & $\pm$0.003 & 27.048 & $\pm$0.004 \\
     & GB-LSR-Scalar-ASR-nf96+noLE & 36.355 & $\pm$0.016 & 32.676 & $\pm$0.010 & 30.696 & $\pm$0.007 & 28.420 & $\pm$0.007 & 27.061 & $\pm$0.003 \\
     & LIIF-RDN & 36.504 &  & 32.748 &  & 30.753 &  & 28.467 &  & 27.096 &  \\
     & LTE-RDN & 36.557 &  & 32.807 &  & 30.808 &  & 28.524 &  & 27.156 &  \\
     & SRNO-RDN & 36.664 &  & 32.898 &  & 30.898 &  & 28.591 &  & 27.215 &  \\
     & OPE-SR-RDN & 36.021 &  & 32.659 &  & 30.734 &  & 28.456 &  & 27.095 &  \\
     & LTE-SwinIR & \textbf{36.756} &  & \textbf{32.978} &  & \textbf{30.981} &  & \textbf{28.672} &  & \textbf{27.292} &  \\
     & LMF & 36.173 &  & 32.443 &  & 30.473 &  & 28.221 &  & 26.872 &  \\
     & SRNO-EDSR & 36.334 &  & 32.572 &  & 30.614 &  & 28.359 &  & 26.997 &  \\
     & OPE-SR-EDSR & 35.835 &  & 32.424 &  & 30.498 &  & 28.254 &  & 26.913 &  \\
     & GSASR & 36.651 &  & 32.892 &  & 30.891 &  & 28.601 &  & 27.223 &  \\
     & Thera & 36.647 &  & 32.931 &  & 30.926 &  & 28.621 &  & 27.244 &  \\
    \bottomrule
  \end{tabular}
\end{table}

%% file: tables/asr_grid_ssim_appendix.tex
\begin{table}[p]
  \centering
  \caption{\textbf{Arbitrary-scale super-resolution quality from $\times 2$ to $\times 8$: SSIM-Y, higher is better.} Scored as in Section~\ref{sec:experiments}; DIV2Kval is the DIV2K validation set. The GB-LSR-Scalar-ASR rows give the mean and sample standard deviation over three training seeds; the other rows are single checkpoints released by their authors. OPE-SR-RDN and OPE-SR-EDSR were trained at $\times 4$ only. Thera is scored without its self-ensemble. Bold: best value in each column of each test set. Row names, encoders, and parameter counts follow Table~\ref{tab:asr_quality}.}
  \label{tab:asr_grid_ssim}
  \scriptsize
  \setlength{\tabcolsep}{2.5pt}
  \begin{tabular}{llr@{\,}lr@{\,}lr@{\,}lr@{\,}lr@{\,}l}
    \toprule
    Test set & Method & \multicolumn{2}{c}{$\times 2$} & \multicolumn{2}{c}{$\times 3$} & \multicolumn{2}{c}{$\times 4$} & \multicolumn{2}{c}{$\times 6$} & \multicolumn{2}{c}{$\times 8$} \\
    \midrule
    Set5 & GB-LSR-Scalar-ASR (base) & 0.9657 & $\pm$0.0000 & 0.9360 & $\pm$0.0002 & 0.9058 & $\pm$0.0005 & 0.8330 & $\pm$0.0013 & 0.7663 & $\pm$0.0018 \\
     & GB-LSR-Scalar-ASR-noLE & 0.9657 & $\pm$0.0001 & 0.9360 & $\pm$0.0004 & 0.9058 & $\pm$0.0006 & 0.8335 & $\pm$0.0009 & 0.7671 & $\pm$0.0022 \\
     & GB-LSR-Scalar-ASR-nf96+noLE & 0.9657 & $\pm$0.0001 & 0.9361 & $\pm$0.0003 & 0.9058 & $\pm$0.0003 & 0.8335 & $\pm$0.0011 & 0.7672 & $\pm$0.0011 \\
     & LIIF-RDN & 0.9660 &  & 0.9380 &  & 0.9091 &  & 0.8418 &  & 0.7803 &  \\
     & LTE-RDN & 0.9662 &  & 0.9384 &  & 0.9100 &  & 0.8431 &  & 0.7816 &  \\
     & SRNO-RDN & \textbf{0.9666} &  & 0.9391 &  & 0.9110 &  & 0.8458 &  & 0.7841 &  \\
     & OPE-SR-RDN & 0.9644 &  & 0.9373 &  & 0.9086 &  & 0.8409 &  & 0.7790 &  \\
     & LTE-SwinIR & \textbf{0.9666} &  & \textbf{0.9396} &  & \textbf{0.9126} &  & \textbf{0.8489} &  & \textbf{0.7849} &  \\
     & LMF & 0.9653 &  & 0.9360 &  & 0.9058 &  & 0.8351 &  & 0.7711 &  \\
     & SRNO-EDSR & 0.9659 &  & 0.9368 &  & 0.9076 &  & 0.8391 &  & 0.7762 &  \\
     & OPE-SR-EDSR & 0.9637 &  & 0.9354 &  & 0.9048 &  & 0.8336 &  & 0.7674 &  \\
     & GSASR & \textbf{0.9666} &  & 0.9393 &  & 0.9123 &  & 0.8473 &  & 0.7843 &  \\
     & Thera & 0.9663 &  & 0.9388 &  & 0.9119 &  & 0.8476 &  & 0.7815 &  \\
    \midrule
    Set14 & GB-LSR-Scalar-ASR (base) & 0.9284 & $\pm$0.0002 & 0.8611 & $\pm$0.0001 & 0.8037 & $\pm$0.0001 & 0.7106 & $\pm$0.0004 & 0.6454 & $\pm$0.0005 \\
     & GB-LSR-Scalar-ASR-noLE & 0.9285 & $\pm$0.0004 & 0.8607 & $\pm$0.0003 & 0.8036 & $\pm$0.0001 & 0.7107 & $\pm$0.0002 & 0.6457 & $\pm$0.0005 \\
     & GB-LSR-Scalar-ASR-nf96+noLE & 0.9285 & $\pm$0.0001 & 0.8608 & $\pm$0.0003 & 0.8039 & $\pm$0.0001 & 0.7108 & $\pm$0.0001 & 0.6460 & $\pm$0.0005 \\
     & LIIF-RDN & 0.9289 &  & 0.8622 &  & 0.8046 &  & 0.7141 &  & 0.6496 &  \\
     & LTE-RDN & 0.9292 &  & 0.8625 &  & 0.8059 &  & 0.7153 &  & 0.6510 &  \\
     & SRNO-RDN & 0.9311 &  & 0.8643 &  & 0.8078 &  & 0.7181 &  & 0.6547 &  \\
     & OPE-SR-RDN & 0.9261 &  & 0.8622 &  & 0.8042 &  & 0.7140 &  & 0.6499 &  \\
     & LTE-SwinIR & \textbf{0.9312} &  & \textbf{0.8661} &  & 0.8094 &  & \textbf{0.7197} &  & \textbf{0.6566} &  \\
     & LMF & 0.9263 &  & 0.8586 &  & 0.7997 &  & 0.7072 &  & 0.6408 &  \\
     & SRNO-EDSR & 0.9282 &  & 0.8602 &  & 0.8030 &  & 0.7104 &  & 0.6461 &  \\
     & OPE-SR-EDSR & 0.9247 &  & 0.8586 &  & 0.8000 &  & 0.7069 &  & 0.6417 &  \\
     & GSASR & 0.9298 &  & 0.8641 &  & 0.8072 &  & 0.7173 &  & 0.6543 &  \\
     & Thera & 0.9307 &  & 0.8659 &  & \textbf{0.8095} &  & 0.7191 &  & 0.6552 &  \\
    \midrule
    B100 & GB-LSR-Scalar-ASR (base) & 0.9110 & $\pm$0.0001 & 0.8261 & $\pm$0.0001 & 0.7606 & $\pm$0.0002 & 0.6648 & $\pm$0.0002 & 0.6029 & $\pm$0.0001 \\
     & GB-LSR-Scalar-ASR-noLE & 0.9109 & $\pm$0.0001 & 0.8254 & $\pm$0.0002 & 0.7606 & $\pm$0.0002 & 0.6647 & $\pm$0.0002 & 0.6031 & $\pm$0.0001 \\
     & GB-LSR-Scalar-ASR-nf96+noLE & 0.9110 & $\pm$0.0000 & 0.8252 & $\pm$0.0003 & 0.7606 & $\pm$0.0002 & 0.6649 & $\pm$0.0004 & 0.6036 & $\pm$0.0004 \\
     & LIIF-RDN & 0.9116 &  & 0.8291 &  & 0.7629 &  & 0.6677 &  & 0.6061 &  \\
     & LTE-RDN & 0.9120 &  & 0.8295 &  & 0.7637 &  & 0.6683 &  & 0.6066 &  \\
     & SRNO-RDN & 0.9130 &  & 0.8306 &  & 0.7654 &  & 0.6704 &  & 0.6094 &  \\
     & OPE-SR-RDN & 0.9096 &  & 0.8285 &  & 0.7617 &  & 0.6663 &  & 0.6049 &  \\
     & LTE-SwinIR & 0.9130 &  & 0.8319 &  & 0.7673 &  & 0.6728 &  & \textbf{0.6116} &  \\
     & LMF & 0.9097 &  & 0.8255 &  & 0.7580 &  & 0.6605 &  & 0.5986 &  \\
     & SRNO-EDSR & 0.9112 &  & 0.8269 &  & 0.7610 &  & 0.6649 &  & 0.6041 &  \\
     & OPE-SR-EDSR & 0.9082 &  & 0.8254 &  & 0.7574 &  & 0.6602 &  & 0.5984 &  \\
     & GSASR & 0.9126 &  & 0.8310 &  & 0.7664 &  & 0.6721 &  & 0.6109 &  \\
     & Thera & \textbf{0.9131} &  & \textbf{0.8320} &  & \textbf{0.7676} &  & \textbf{0.6732} &  & 0.6110 &  \\
    \midrule
    Urban100 & GB-LSR-Scalar-ASR (base) & 0.9380 & $\pm$0.0005 & 0.8731 & $\pm$0.0004 & 0.8110 & $\pm$0.0004 & 0.7052 & $\pm$0.0004 & 0.6296 & $\pm$0.0006 \\
     & GB-LSR-Scalar-ASR-noLE & 0.9380 & $\pm$0.0003 & 0.8728 & $\pm$0.0001 & 0.8111 & $\pm$0.0001 & 0.7049 & $\pm$0.0004 & 0.6295 & $\pm$0.0005 \\
     & GB-LSR-Scalar-ASR-nf96+noLE & 0.9385 & $\pm$0.0001 & 0.8729 & $\pm$0.0003 & 0.8113 & $\pm$0.0007 & 0.7054 & $\pm$0.0005 & 0.6305 & $\pm$0.0002 \\
     & LIIF-RDN & 0.9405 &  & 0.8770 &  & 0.8174 &  & 0.7139 &  & 0.6387 &  \\
     & LTE-RDN & 0.9417 &  & 0.8791 &  & 0.8205 &  & 0.7166 &  & 0.6412 &  \\
     & SRNO-RDN & 0.9440 &  & 0.8816 &  & 0.8248 &  & 0.7245 &  & 0.6506 &  \\
     & OPE-SR-RDN & 0.9338 &  & 0.8739 &  & 0.8136 &  & 0.7075 &  & 0.6320 &  \\
     & LTE-SwinIR & \textbf{0.9454} &  & \textbf{0.8868} &  & \textbf{0.8322} &  & \textbf{0.7320} &  & \textbf{0.6569} &  \\
     & LMF & 0.9340 &  & 0.8652 &  & 0.8008 &  & 0.6934 &  & 0.6176 &  \\
     & SRNO-EDSR & 0.9379 &  & 0.8709 &  & 0.8110 &  & 0.7072 &  & 0.6324 &  \\
     & OPE-SR-EDSR & 0.9298 &  & 0.8653 &  & 0.8010 &  & 0.6928 &  & 0.6173 &  \\
     & GSASR & 0.9438 &  & 0.8832 &  & 0.8267 &  & 0.7273 &  & 0.6531 &  \\
     & Thera & 0.9433 &  & 0.8846 &  & 0.8277 &  & 0.7277 &  & 0.6522 &  \\
    \midrule
    DIV2Kval & GB-LSR-Scalar-ASR (base) & 0.9523 & $\pm$0.0001 & 0.9026 & $\pm$0.0001 & 0.8557 & $\pm$0.0001 & 0.7790 & $\pm$0.0002 & 0.7251 & $\pm$0.0002 \\
     & GB-LSR-Scalar-ASR-noLE & 0.9522 & $\pm$0.0000 & 0.9024 & $\pm$0.0000 & 0.8557 & $\pm$0.0001 & 0.7789 & $\pm$0.0001 & 0.7252 & $\pm$0.0002 \\
     & GB-LSR-Scalar-ASR-nf96+noLE & 0.9524 & $\pm$0.0000 & 0.9024 & $\pm$0.0002 & 0.8559 & $\pm$0.0002 & 0.7793 & $\pm$0.0002 & 0.7256 & $\pm$0.0002 \\
     & LIIF-RDN & 0.9531 &  & 0.9037 &  & 0.8572 &  & 0.7813 &  & 0.7279 &  \\
     & LTE-RDN & 0.9534 &  & 0.9043 &  & 0.8581 &  & 0.7824 &  & 0.7292 &  \\
     & SRNO-RDN & 0.9542 &  & 0.9054 &  & 0.8599 &  & 0.7850 &  & 0.7320 &  \\
     & OPE-SR-RDN & 0.9510 &  & 0.9031 &  & 0.8565 &  & 0.7804 &  & 0.7269 &  \\
     & LTE-SwinIR & \textbf{0.9548} &  & \textbf{0.9069} &  & \textbf{0.8617} &  & \textbf{0.7871} &  & \textbf{0.7343} &  \\
     & LMF & 0.9509 &  & 0.8994 &  & 0.8509 &  & 0.7732 &  & 0.7193 &  \\
     & SRNO-EDSR & 0.9520 &  & 0.9009 &  & 0.8539 &  & 0.7777 &  & 0.7245 &  \\
     & OPE-SR-EDSR & 0.9495 &  & 0.8996 &  & 0.8512 &  & 0.7737 &  & 0.7199 &  \\
     & GSASR & 0.9542 &  & 0.9059 &  & 0.8605 &  & 0.7858 &  & 0.7328 &  \\
     & Thera & 0.9542 &  & 0.9065 &  & 0.8613 &  & 0.7865 &  & 0.7331 &  \\
    \bottomrule
  \end{tabular}
\end{table}

%% file: tables/asr_grid_lpips_appendix.tex
\begin{table}[p]
  \centering
  \caption{\textbf{Arbitrary-scale super-resolution quality from $\times 2$ to $\times 8$: LPIPS, lower is better.} Scored as in Section~\ref{sec:experiments}; DIV2Kval is the DIV2K validation set. The GB-LSR-Scalar-ASR rows give the mean and sample standard deviation over three training seeds; the other rows are single checkpoints released by their authors. OPE-SR-RDN and OPE-SR-EDSR were trained at $\times 4$ only. Thera is scored without its self-ensemble. Bold: best value in each column of each test set. Row names, encoders, and parameter counts follow Table~\ref{tab:asr_quality}.}
  \label{tab:asr_grid_lpips}
  \scriptsize
  \setlength{\tabcolsep}{2.5pt}
  \begin{tabular}{llr@{\,}lr@{\,}lr@{\,}lr@{\,}lr@{\,}l}
    \toprule
    Test set & Method & \multicolumn{2}{c}{$\times 2$} & \multicolumn{2}{c}{$\times 3$} & \multicolumn{2}{c}{$\times 4$} & \multicolumn{2}{c}{$\times 6$} & \multicolumn{2}{c}{$\times 8$} \\
    \midrule
    Set5 & GB-LSR-Scalar-ASR (base) & 0.0572 & $\pm$0.0005 & 0.1333 & $\pm$0.0009 & 0.1891 & $\pm$0.0011 & 0.2816 & $\pm$0.0028 & 0.3637 & $\pm$0.0043 \\
     & GB-LSR-Scalar-ASR-noLE & 0.0569 & $\pm$0.0004 & 0.1329 & $\pm$0.0009 & 0.1886 & $\pm$0.0016 & 0.2795 & $\pm$0.0022 & 0.3593 & $\pm$0.0052 \\
     & GB-LSR-Scalar-ASR-nf96+noLE & 0.0571 & $\pm$0.0007 & 0.1333 & $\pm$0.0011 & 0.1900 & $\pm$0.0013 & 0.2801 & $\pm$0.0041 & 0.3612 & $\pm$0.0027 \\
     & LIIF-RDN & 0.0559 &  & 0.1235 &  & 0.1732 &  & 0.2399 &  & 0.3006 &  \\
     & LTE-RDN & 0.0553 &  & 0.1229 &  & 0.1745 &  & 0.2529 &  & 0.3182 &  \\
     & SRNO-RDN & 0.0555 &  & 0.1214 &  & 0.1730 &  & 0.2421 &  & 0.2995 &  \\
     & OPE-SR-RDN & \textbf{0.0528} &  & 0.1218 &  & 0.1740 &  & 0.2508 &  & 0.3171 &  \\
     & LTE-SwinIR & 0.0553 &  & 0.1212 &  & 0.1725 &  & 0.2434 &  & 0.3106 &  \\
     & LMF & 0.0576 &  & 0.1256 &  & 0.1777 &  & 0.2437 &  & 0.3033 &  \\
     & SRNO-EDSR & 0.0567 &  & 0.1235 &  & 0.1773 &  & 0.2475 &  & 0.3090 &  \\
     & OPE-SR-EDSR & 0.0567 &  & 0.1251 &  & 0.1784 &  & 0.2619 &  & 0.3391 &  \\
     & GSASR & 0.0533 &  & \textbf{0.1203} &  & \textbf{0.1724} &  & \textbf{0.2382} &  & \textbf{0.2931} &  \\
     & Thera & 0.0546 &  & 0.1246 &  & 0.1748 &  & 0.2467 &  & 0.3088 &  \\
    \midrule
    Set14 & GB-LSR-Scalar-ASR (base) & 0.0888 & $\pm$0.0005 & 0.2056 & $\pm$0.0005 & 0.2906 & $\pm$0.0009 & 0.4068 & $\pm$0.0011 & 0.4766 & $\pm$0.0018 \\
     & GB-LSR-Scalar-ASR-noLE & 0.0892 & $\pm$0.0006 & 0.2071 & $\pm$0.0005 & 0.2903 & $\pm$0.0014 & 0.4059 & $\pm$0.0010 & 0.4762 & $\pm$0.0018 \\
     & GB-LSR-Scalar-ASR-nf96+noLE & 0.0888 & $\pm$0.0003 & 0.2059 & $\pm$0.0014 & 0.2905 & $\pm$0.0007 & 0.4063 & $\pm$0.0013 & 0.4773 & $\pm$0.0021 \\
     & LIIF-RDN & 0.0888 &  & 0.1974 &  & 0.2792 &  & 0.3791 &  & 0.4410 &  \\
     & LTE-RDN & 0.0885 &  & 0.1986 &  & 0.2790 &  & 0.3892 &  & 0.4577 &  \\
     & SRNO-RDN & 0.0876 &  & 0.1978 &  & 0.2763 &  & 0.3784 &  & \textbf{0.4404} &  \\
     & OPE-SR-RDN & 0.0867 &  & 0.1993 &  & 0.2827 &  & 0.3929 &  & 0.4625 &  \\
     & LTE-SwinIR & 0.0866 &  & 0.1944 &  & 0.2713 &  & 0.3823 &  & 0.4519 &  \\
     & LMF & 0.0957 &  & 0.2052 &  & 0.2873 &  & 0.3859 &  & 0.4456 &  \\
     & SRNO-EDSR & 0.0920 &  & 0.2029 &  & 0.2812 &  & 0.3875 &  & 0.4466 &  \\
     & OPE-SR-EDSR & 0.0929 &  & 0.2059 &  & 0.2902 &  & 0.4043 &  & 0.4760 &  \\
     & GSASR & \textbf{0.0841} &  & \textbf{0.1917} &  & \textbf{0.2706} &  & \textbf{0.3778} &  & 0.4411 &  \\
     & Thera & 0.0844 &  & 0.1950 &  & 0.2746 &  & 0.3825 &  & 0.4468 &  \\
    \midrule
    B100 & GB-LSR-Scalar-ASR (base) & 0.1520 & $\pm$0.0017 & 0.3045 & $\pm$0.0013 & 0.3893 & $\pm$0.0014 & 0.5031 & $\pm$0.0008 & 0.5753 & $\pm$0.0008 \\
     & GB-LSR-Scalar-ASR-noLE & 0.1529 & $\pm$0.0010 & 0.3049 & $\pm$0.0006 & 0.3887 & $\pm$0.0013 & 0.5027 & $\pm$0.0002 & 0.5748 & $\pm$0.0001 \\
     & GB-LSR-Scalar-ASR-nf96+noLE & 0.1527 & $\pm$0.0014 & 0.3048 & $\pm$0.0027 & 0.3903 & $\pm$0.0014 & 0.5039 & $\pm$0.0025 & 0.5753 & $\pm$0.0022 \\
     & LIIF-RDN & 0.1439 &  & 0.2756 &  & 0.3629 &  & 0.4749 &  & 0.5415 &  \\
     & LTE-RDN & 0.1417 &  & 0.2725 &  & 0.3614 &  & 0.4845 &  & 0.5587 &  \\
     & SRNO-RDN & 0.1370 &  & 0.2693 &  & 0.3558 &  & \textbf{0.4698} &  & \textbf{0.5383} &  \\
     & OPE-SR-RDN & \textbf{0.1350} &  & 0.2709 &  & 0.3655 &  & 0.4908 &  & 0.5643 &  \\
     & LTE-SwinIR & 0.1383 &  & \textbf{0.2673} &  & \textbf{0.3547} &  & 0.4759 &  & 0.5496 &  \\
     & LMF & 0.1496 &  & 0.2825 &  & 0.3719 &  & 0.4823 &  & 0.5474 &  \\
     & SRNO-EDSR & 0.1436 &  & 0.2770 &  & 0.3618 &  & 0.4775 &  & 0.5453 &  \\
     & OPE-SR-EDSR & 0.1397 &  & 0.2788 &  & 0.3756 &  & 0.5030 &  & 0.5773 &  \\
     & GSASR & 0.1356 &  & 0.2681 &  & 0.3581 &  & 0.4716 &  & 0.5417 &  \\
     & Thera & 0.1407 &  & 0.2834 &  & 0.3664 &  & 0.4774 &  & 0.5474 &  \\
    \midrule
    Urban100 & GB-LSR-Scalar-ASR (base) & 0.0591 & $\pm$0.0008 & 0.1534 & $\pm$0.0003 & 0.2334 & $\pm$0.0016 & 0.3544 & $\pm$0.0017 & 0.4446 & $\pm$0.0016 \\
     & GB-LSR-Scalar-ASR-noLE & 0.0590 & $\pm$0.0006 & 0.1531 & $\pm$0.0003 & 0.2329 & $\pm$0.0014 & 0.3545 & $\pm$0.0006 & 0.4456 & $\pm$0.0009 \\
     & GB-LSR-Scalar-ASR-nf96+noLE & 0.0587 & $\pm$0.0002 & 0.1534 & $\pm$0.0022 & 0.2341 & $\pm$0.0030 & 0.3552 & $\pm$0.0033 & 0.4460 & $\pm$0.0037 \\
     & LIIF-RDN & 0.0569 &  & 0.1387 &  & 0.2088 &  & 0.3114 &  & 0.3920 &  \\
     & LTE-RDN & 0.0553 &  & 0.1379 &  & 0.2059 &  & 0.3233 &  & 0.4127 &  \\
     & SRNO-RDN & 0.0521 &  & 0.1337 &  & 0.2000 &  & \textbf{0.2980} &  & \textbf{0.3734} &  \\
     & OPE-SR-RDN & 0.0563 &  & 0.1407 &  & 0.2152 &  & 0.3384 &  & 0.4284 &  \\
     & LTE-SwinIR & \textbf{0.0516} &  & \textbf{0.1294} &  & \textbf{0.1939} &  & 0.3089 &  & 0.3952 &  \\
     & LMF & 0.0663 &  & 0.1570 &  & 0.2314 &  & 0.3357 &  & 0.4161 &  \\
     & SRNO-EDSR & 0.0599 &  & 0.1474 &  & 0.2148 &  & 0.3160 &  & 0.3954 &  \\
     & OPE-SR-EDSR & 0.0628 &  & 0.1544 &  & 0.2338 &  & 0.3602 &  & 0.4541 &  \\
     & GSASR & 0.0525 &  & 0.1323 &  & 0.1998 &  & 0.2993 &  & 0.3776 &  \\
     & Thera & 0.0522 &  & 0.1352 &  & 0.2081 &  & 0.3200 &  & 0.4025 &  \\
    \midrule
    DIV2Kval & GB-LSR-Scalar-ASR (base) & 0.0859 & $\pm$0.0009 & 0.1967 & $\pm$0.0007 & 0.2725 & $\pm$0.0009 & 0.3721 & $\pm$0.0006 & 0.4388 & $\pm$0.0009 \\
     & GB-LSR-Scalar-ASR-noLE & 0.0857 & $\pm$0.0006 & 0.1964 & $\pm$0.0007 & 0.2719 & $\pm$0.0002 & 0.3717 & $\pm$0.0005 & 0.4387 & $\pm$0.0003 \\
     & GB-LSR-Scalar-ASR-nf96+noLE & 0.0861 & $\pm$0.0010 & 0.1966 & $\pm$0.0033 & 0.2727 & $\pm$0.0018 & 0.3722 & $\pm$0.0023 & 0.4390 & $\pm$0.0024 \\
     & LIIF-RDN & 0.0836 &  & 0.1864 &  & 0.2597 &  & 0.3505 &  & 0.4109 &  \\
     & LTE-RDN & 0.0826 &  & 0.1859 &  & 0.2590 &  & 0.3585 &  & 0.4247 &  \\
     & SRNO-RDN & 0.0796 &  & 0.1841 &  & 0.2556 &  & \textbf{0.3476} &  & \textbf{0.4083} &  \\
     & OPE-SR-RDN & 0.0817 &  & 0.1878 &  & 0.2632 &  & 0.3649 &  & 0.4319 &  \\
     & LTE-SwinIR & 0.0790 &  & 0.1811 &  & \textbf{0.2530} &  & 0.3517 &  & 0.4168 &  \\
     & LMF & 0.0903 &  & 0.1971 &  & 0.2712 &  & 0.3612 &  & 0.4206 &  \\
     & SRNO-EDSR & 0.0857 &  & 0.1928 &  & 0.2647 &  & 0.3586 &  & 0.4201 &  \\
     & OPE-SR-EDSR & 0.0870 &  & 0.1966 &  & 0.2739 &  & 0.3776 &  & 0.4455 &  \\
     & GSASR & \textbf{0.0774} &  & \textbf{0.1798} &  & 0.2545 &  & 0.3486 &  & 0.4105 &  \\
     & Thera & 0.0798 &  & 0.1861 &  & 0.2597 &  & 0.3545 &  & 0.4166 &  \\
    \bottomrule
  \end{tabular}
\end{table}

%% file: tables/asr_ood_ladder_appendix.tex
\begin{table}[t]
  \centering
  \caption{\textbf{PSNR-Y at large out-of-distribution scales.} DIV2K validation split at $\times 12$ to $\times 30$, scored as in Section~\ref{sec:experiments} and therefore not comparable with published DIV2K numbers. GB-LSR-Scalar-ASR (base) is the mean over three training seeds, with a seed standard deviation of at most 0.006\,dB at every scale; the other rows are single checkpoints released by their authors. Params: trainable parameters in millions. OPE-SR-RDN was trained at $\times 4$ only. SRNO-RDN and the five recent methods on EDSR-baseline encoders were not scored at these scales.}
  \label{tab:asr_ood_ladder}
  \footnotesize
  \setlength{\tabcolsep}{4pt}
  \begin{tabular}{lrrrrr}
    \toprule
    Method & Params (M) & $\times 12$ & $\times 18$ & $\times 24$ & $\times 30$ \\
    \midrule
    GB-LSR-Scalar-ASR (base) & 22.02 & 25.371 & 23.840 & 22.830 & 22.121 \\
    \midrule
    LIIF-RDN & 22.32 & 25.394 & 23.854 & 22.828 & 22.115 \\
    LTE-RDN & 22.47 & 25.456 & 23.916 & 22.882 & 22.167 \\
    LTE-SwinIR & 12.10 & 25.584 & 24.012 & 22.981 & 22.248 \\
    OPE-SR-RDN & 29.65 & 25.416 & 23.883 & 22.860 & 22.152 \\
    \bottomrule
  \end{tabular}
\end{table}

%% file: tables/asr_sensitivity_appendix.tex
\begin{table}[t]
  \centering
  \caption{\textbf{Cutoff and bandwidth initialization of the super-resolution decoder.} GB-LSR-Scalar-ASR (base, first row: $p_{\max} = 16$, bandwidth initialized at $s = 1.0$) and variants that change one of the two, three seeds each, scored as in Section~\ref{sec:experiments}. PSNR-Y (dB) is averaged over the $15$ in-distribution (ID) cells and the $10$ out-of-distribution (OOD) cells; $\Delta$ is the variant minus the base, paired by seed for the initialization variants (mean $\pm$ seed standard deviation of the paired differences) and as a difference of three-seed means for the cutoff variants ($\pm$ its standard error). The seed standard deviation of the base model's mean ID PSNR-Y is 0.015\,dB. ID worst: the in-distribution cell with the lowest $\Delta$. Learned $s$: the three-seed mean of the bandwidth after $200{,}000$ and $1{,}000{,}000$ steps.}
  \label{tab:asr_sensitivity}
  \scriptsize
  \setlength{\tabcolsep}{3pt}
  \begin{tabular}{rrrrrrlrr}
    \toprule
    $p_{\max}$ & Initial $s$ & \multicolumn{2}{c}{ID PSNR-Y (dB)} & \multicolumn{2}{c}{OOD PSNR-Y (dB)} & ID worst & \multicolumn{2}{c}{Learned $s$} \\
    \cmidrule(lr){3-4} \cmidrule(lr){5-6} \cmidrule(lr){8-9}
     &  & mean & $\Delta$ & mean & $\Delta$ &  & $200$k & $1$M \\
    \midrule
    16 & 1.0 & 31.599 &  & 26.028 &  &  & 0.916 & 0.880 \\
    \midrule
    12 & 1.0 & 31.593 & $-0.006 \pm 0.011$ & 26.026 & $-0.002 \pm 0.012$ & Urban100 $\times 2$, $-0.044$ & 0.892 & 0.855 \\
    20 & 1.0 & 31.599 & $0.000 \pm 0.011$ & 26.023 & $-0.005 \pm 0.004$ & Set5 $\times 4$, $-0.041$ & 0.932 & 0.901 \\
    24 & 1.0 & 31.605 & $+0.006 \pm 0.010$ & 26.035 & $+0.007 \pm 0.007$ & DIV2Kval $\times 3$, $-0.016$ & 0.942 & 0.911 \\
    16 & 0.5 & 31.595 & $-0.004 \pm 0.027$ & 26.021 & $-0.007 \pm 0.014$ & Urban100 $\times 4$, $-0.035$ & 0.478 & 0.481 \\
    16 & 2.0 & 29.373 & $-2.226 \pm 0.029$ & 24.231 & $-1.798 \pm 0.001$ & Urban100 $\times 3$, $-3.712$ & 1.966 & 1.935 \\
    \bottomrule
  \end{tabular}
\end{table}

%% file: tables/cell_boundary_appendix.tex
\begin{table}[t]
  \centering
  \caption{\textbf{Continuity at low-resolution cell boundaries.} Luma in gray levels out of $255$ on all $319$ images of the five benchmarks. $J$: mean absolute jump across a cell boundary at an offset of $10^{-4}$ cell; ratio: $J(10^{-3})/J(10^{-4})$, about $10$ for an output that is continuous in value and about $1$ for a jump; $E$: boundary excess of adjacent-pixel differences, $1$ for none; band gap: PSNR-Y away from cell boundaries minus next to them; PSNR-Y: whole image. Base and No ensemble are three-seed means of the base variant and of the variant trained without the local ensemble, followed by their difference and its $95\%$ bootstrap interval. Off at test is one of the three base models with the ensemble disabled at test time, a positive control; Bicubic is cubic convolution with $a = -0.75$; Nearest is nearest-neighbor interpolation, for which $E$ is undefined because its differences vanish within cells. The native rows give $E$ at patch boundaries on the $138$ held-out native images, with its $95\%$ bootstrap interval.}
  \label{tab:cell_boundary}
  \scriptsize
  \setlength{\tabcolsep}{3pt}
  \begin{tabular}{llrrrrr}
    \toprule
    Scale & Row & $J$ (gray) & Ratio & $E$ & Band gap (dB) & PSNR-Y (dB) \\
    \midrule
    $\times 4$ & Base & 0.0035 & 10.0 & 1.040 & 0.691 & 28.350 \\
     & No ensemble & 0.225 & 1.04 & 1.041 & 0.688 & 28.352 \\
     & Difference & $+0.221$ &  & $+0.002$ & $-0.002$ & $+0.002$ \\
     &  & [$+0.208$, $+0.235$] &  & [$-0.002$, $+0.006$] & [$-0.007$, $+0.001$] & [$-0.012$, $+0.016$] \\
     & Bicubic & 0.0023 & 10.0 & 1.018 & 0.840 & 25.893 \\
     & Nearest & 10.75 & 1.00 &  & 2.027 & 24.667 \\
     & Off at test & 26.19 & 1.00 & 3.714 & 4.765 & 23.135 \\
    \midrule
    $\times 8$ & Base & 0.0043 & 10.0 & 1.006 & 0.321 & 24.904 \\
     & No ensemble & 0.312 & 1.04 & 1.016 & 0.316 & 24.901 \\
     & Difference & $+0.307$ &  & $+0.010$ & $-0.004$ & $-0.002$ \\
     &  & [$+0.290$, $+0.326$] &  & [$+0.006$, $+0.013$] & [$-0.007$, $-0.002$] & [$-0.012$, $+0.006$] \\
     & Bicubic & 0.0028 & 10.0 & 1.003 & 0.425 & 23.279 \\
     & Nearest & 12.67 & 1.00 &  & 1.265 & 22.332 \\
     & Off at test & 30.94 & 1.00 & 10.868 & 2.377 & 21.095 \\
    \midrule
    Native & $P = 16$ &  &  & 1.217 &  &  \\
     &  &  &  & [1.198, 1.235] &  &  \\
     & $P = 32$ &  &  & 2.045 &  &  \\
     &  &  &  & [1.955, 2.142] &  &  \\
    \bottomrule
  \end{tabular}
\end{table}

%% file: tables/compute_total.tex
Summing the runs in Tables~\ref{tab:compute_native}--\ref{tab:compute_asr} gives $2{,}506.54$ GPU-hours over $99$ runs: $2{,}287.41$~h on H200 ($83$ runs) and $219.13$~h on H100 ($16$ runs).